\documentclass{article}

\usepackage{subfig}    
\usepackage{multirow}   
\usepackage{graphicx}   
\usepackage{float}      

\usepackage{hyperref}

\usepackage{subcaption}    

 \usepackage[accepted]{icml2025}
\usepackage{multirow}
\usepackage{amsmath}
\usepackage{amssymb}
\usepackage{mathtools}
\usepackage{amsthm}
\usepackage{booktabs}
\usepackage[capitalize,noabbrev]{cleveref}

\theoremstyle{plain}
\newtheorem{theorem}{Theorem}[section]

\theoremstyle{definition}

\theoremstyle{remark}

\usepackage[textsize=tiny]{todonotes}

\icmltitlerunning{Submission and Formatting Instructions for ICML 2025}

\begin{document}

\twocolumn[
\icmltitle{Label Distribution Learning with Biased Annotations by Learning  Multi-Label Representation}



\icmlsetsymbol{equal}{*}

\begin{icmlauthorlist}
	\icmlauthor{Zhiqiang Kou}{equal,1}
	\icmlauthor{Si Qin}{equal,1}
	\icmlauthor{Hailin Wang}{equal,5}
	\icmlauthor{Mingkun Xie}{2}
	\icmlauthor{Shuo Chen}{2}
	\icmlauthor{Yuheng Jia}{1}
	\icmlauthor{Tongliang Liu}{4}
	\icmlauthor{Masashi Sugiyama}{2,3}
	\icmlauthor{Xin Geng}{1}
\end{icmlauthorlist}

\icmlaffiliation{1}{Southeast University, China}
\icmlaffiliation{2}{Center for Advanced Intelligence Project, RIKEN, Japan}
\icmlaffiliation{3}{The University of Tokyo, Tokyo, Japan}
\icmlaffiliation{4}{The University of Sydney, Australia}
\icmlaffiliation{5}{Xi'an Jiaotong University, China}

\icmlcorrespondingauthor{Xin Geng}{xgeng@seu.edu.cn}
\icmlkeywords{Machine Learning, ICML}

\vskip 0.3in
]



\printAffiliationsAndNotice{\icmlEqualContribution} 

\begin{abstract}
	
Multi-label learning (MLL) has gained attention for its ability to represent real-world data. Label Distribution Learning (LDL), an extension of MLL to learning from label distributions, faces challenges in collecting accurate label distributions. To address the issue of biased annotations, based on the low-rank assumption, existing works recover true distributions from biased observations by exploring the label correlations. However, recent evidence shows that the label distribution tends to be full-rank, and naive apply of low-rank approximation on biased observation leads to inaccurate recovery and performance degradation. In this paper, we address the LDL with biased annotations problem from a novel perspective,  where we first degenerate the soft label distribution into a hard multi-hot label and then recover the true label information for each instance. This idea stems from an insight that assigning hard multi-hot labels is often easier than assigning a soft label distribution, and it shows stronger immunity to noise disturbances, leading to smaller label bias. Moreover, assuming that the multi-label space for predicting label distributions is low-rank offers a more reasonable approach to capturing label correlations.  Theoretical analysis and experiments confirm the effectiveness and robustness of our method on real-world datasets.

\end{abstract}

\section{Introduction}

Multi-Label Learning (MLL) \cite{6471714} has gained significant attention due to its ability to associate multiple labels with a single instance, making it widely applicable in tasks such as text classification \cite{liu2017deep}  and image annotation \cite{jing2016multi}.

Label Distribution Learning (LDL)\footnote{LDL is similar to learning from soft labels, but the soft-label formulation focuses on single-label problems (i.e., there is only one true label for each instance), while LDL considers multi-label problems (i.e., each instance can have multiple true labels).} \cite{geng2016label} extends MLL by assigning a real-valued \textit{label description  degree} \cite{jia2019facial}  to each label, offering more detailed supervisory information. Leveraging label correlations is a key strategy in MLL \cite{huang2012multi}, and applying low-rank constraints in the output space is an effective approach to capturing these correlations \cite{zhu2017multi}.  Building on the advancements in label correlation modeling in MLL, a branch of algorithms \cite{jia2018label}\cite{jia2019facial}\cite{kou2023instance} has  extended these techniques to LDL by assuming low-rank structures in label distribution spaces, aiming  to capture label correlations more comprehensively.

Annotating label distributions is inherently challenging and often leads to  \textit{biased label distributions}, where the collected label distributions deviate from the true  distributions due to variations in annotators' expertise or subjective judgments\cite{xie2018partial}.  The existing method \cite{he2024generative} \cite{kou2024inaccurate} \cite{xu2017incomplete} aims to address this issue by leveraging label correlations  modeling the clean label distribution  and training LDL models effectively. For instance, IncomLDL \cite{xu2017incomplete} aims to model the learned label distribution space using low-rank label correlations, thereby completing the missing entries in the label distribution matrix. And in  LRS-LDL\cite{kou2023instance}, the noisy label distribution is modeled as \(\hat{\mathbf{D}} = \mathbf{D} + \mathbf{E}\), where $\mathbf{D}$ is the true label distributions and \(\mathbf{E}\) represents noise. It assumes a low-rank structure on the output space (\(\mathbf{D} = \mathbf{W}\mathbf{X}\)) and sparsity of the noise. 
\begin{figure}[!t]
	\centering
	\includegraphics[width=1\linewidth]{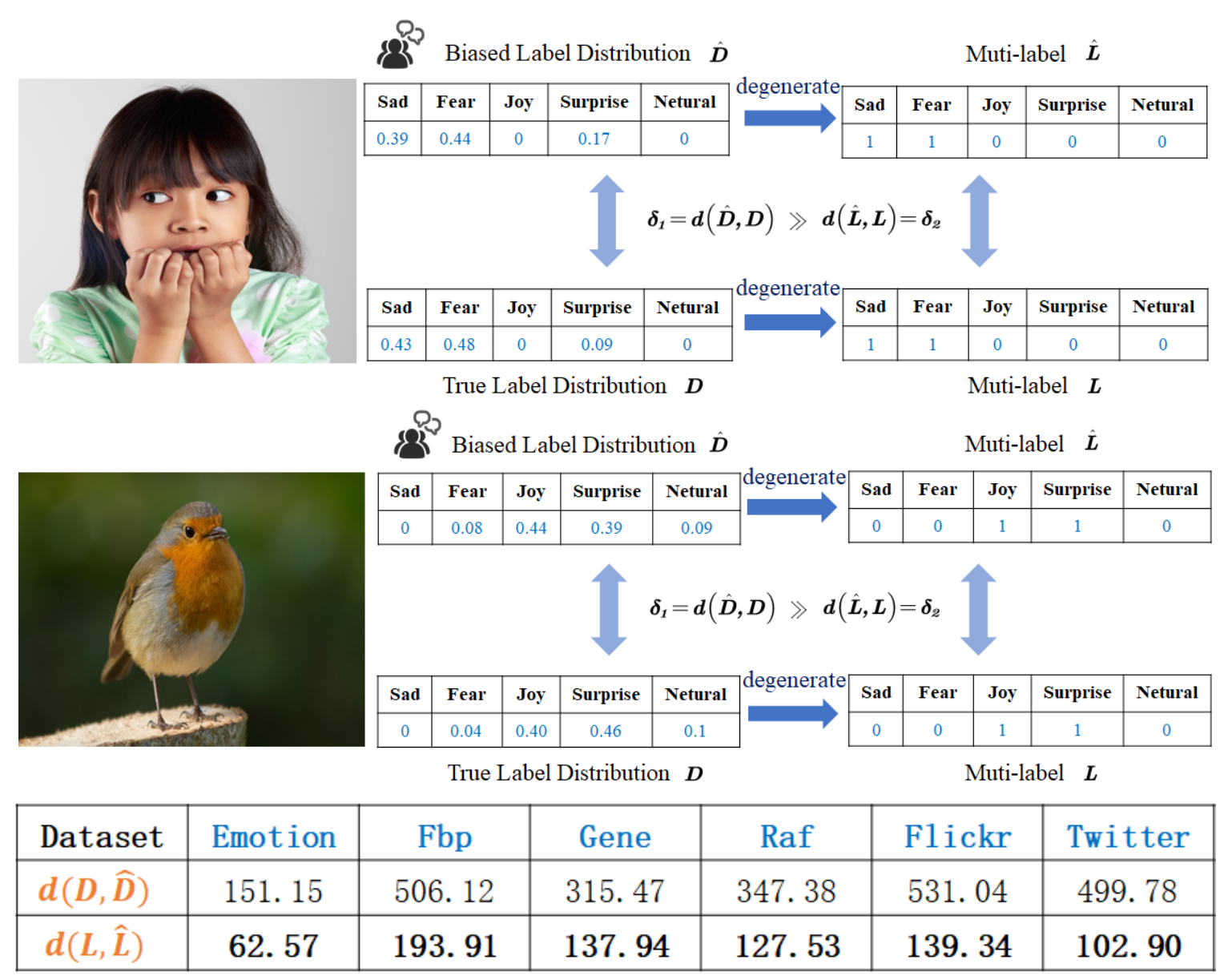}
	\caption{Illustration of biased label distribution learning using examples from the RAF dataset \protect\cite{li2019blended} and the Emotion dataset \protect\cite{peng2015mixed}. Despite discrepancies between the biased label distribution \(\hat{\mathbf{D}}\) and the true distribution \(\mathbf{D}\), their corresponding multi-label representations (\(\hat{\mathbf{L}}\) and \(\mathbf{L}\)) are much closer.}
	\label{fig:fig1}
\end{figure}
Similarly, IDI-LDL \cite{kou2024inaccurate} applies a low-rank assumption on the output space while employing an \(\ell_{2,1}\)-norm constraint on the noise.

However, recent evidence shows that the label distribution tends to be full-rank \cite{wang2021label}, and naive apply of low-rank approximation on biased observation leads to inaccurate recovery and performance degradation.  In this paper, we address the LDL with biased annotations problem from a novel perspective,  where we first degenerate the soft label distribution into a hard multi-hot label and then recover the true label information for each instance.
This idea stems from an insight that assigning hard multi-hot labels is often easier than assigning a soft label distribution, and it shows stronger immunity to noise disturbances, leading to smaller label bias. As shown in Fig. \ref{fig:fig1}, two images from the Real-world Affective Faces (RAF) \cite{li2019blended} and Emotion datasets \cite{peng2015mixed} reveal significant discrepancies between the biased and true soft label distributions (\(\hat{\mathbf{D}}\) and \(\mathbf{D}\)). However, their corresponding multi-hot label (\(\hat{\mathbf{L}}\) and \(\mathbf{L}\)) exhibit much smaller differences. This phenomenon holds across datasets, as demonstrated by the Tables in Fig. \ref{fig:fig1}.  Moreover, while label distributions are inherently full-rank \cite{wang2021label}, multi-label spaces are widely regarded as low-rank \cite{zhu2017multi}, making them more computationally efficient for correlation modeling.   The contributions of this work are summarized as

\begin{itemize}
	\item We utilize multi-label information and label correlations to model the recovery of the true label distribution and propose the BLDL algorithm (Sections 2 and 3).
	\item Extensive theoretical analysis is provided, including convergence guarantees and generalization error bounds (Section 4).
	\item The effectiveness of the method is validated through comprehensive experiments, with superior performance demonstrated and insights verified. (Section 5).
\end{itemize}

\subsection{Related Work}
Label Distribution Learning (LDL) \cite{6475129} is a learning paradigm designed to predict label distributions that capture the degree of association between each instance and its labels. Unlike traditional binary or multi-label learning, LDL provides a more detailed and comprehensive framework by leveraging \textit{label description degrees}, which quantify the relevance of each label to an instance \cite{10275121}. Since its inception, various models have been developed to enhance LDL’s modeling capabilities, such as adapting support vector machines, decision trees, and embedding-based methods, to represent label distributions effectively.

Existing LDL methods assume accurate label distributions, but in real-world scenarios, annotators’ biases, limited expertise, or label familiarity often lead to deviations between annotated and true distributions \cite{10540001}. This motivates the study of \textit{label distribution learning with bias (LDL with biased distributions)}, which focuses on recovering true label distributions while mitigating annotator-induced biases. Our work addresses this challenge by leveraging the hierarchical relationship between biased label distributions and their corresponding multi-label representations.

\textbf{Notations}: Denote \(\mathbf{X} \in \mathbb{R}^{d \times n}\) as the feature matrix, where \(d\) is the feature dimensionality and \(n\) is the number of instances. The label space is \(\mathcal{Y} = \{y_1,\ldots, y_m\}\), where \(m\) is the number of labels. The accurate training set for LDL is \(\mathcal{T} = \{(\mathbf{x}_i, \mathbf{d}_i)\}_{i=1}^n\), with \(\mathbf{d}_i = [d_{\mathbf{x}_i}^{y_1}, \cdots, d_{\mathbf{x}_i}^{y_m}]^\top \in \mathbb{R}^m\), satisfying \(d_{\mathbf{x}_i}^y \geq 0\) for all \(y \in \mathcal{Y}\) and \(\sum_{y} d_{\mathbf{x}_i}^y = 1\). The label distribution matrix is \(\mathbf{D} = [\mathbf{d}_1,\ldots, \mathbf{d}_n] \in \mathbb{R}^{m \times n}\). We assume the observed label distribution \(\hat{\mathbf{D}} \in \mathbb{R}^{m \times n}\) is biased, while the true label distribution is unknown. The goal is to learn a decision function \(\mathfrak{G}: \mathbb{R}^{d \times n} \to \mathbb{R}^{m\times n}\) using the training set \(\{\mathbf{X}, \hat{\mathbf{D}}\}\), such that \(\mathfrak{G}(\mathbf{X}_{i:}) \approx {\mathbf{D}}_{i:}\).

\section{THE BLDL   APPROACH}

A common approach in multi-label learning to capture label correlations is leveraging low-rank modeling on the output space. This can be formulated as the following optimization problem:
\begin{equation}
	\min_{\mathbf{W}} \ \text{rank}(\mathbf{WX}), \quad \text{s.t.} \ \|\mathbf{WX} - \mathbf{L}\|_F \leq \delta,
\end{equation}
where \(\mathbf{W} \in \mathbb{R}^{m \times d}\) is the learned weight matrix mapping the feature space to the multi-label space \(\mathbf{L} \in \mathbb{R}^{m \times n}\). The first term  captures label correlations via low-rank modeling \cite{jia2019facial}.

Annotating label distributions is challenging, often introducing bias. As an extension of MLL, certain bias LDL methods leverage label correlations to recover true distributions from biased ones while learning an effective model. These methods can be formulated as:
\begin{equation}
	\begin{aligned}
		&\min_{\mathbf{W,D,E}} \ \text{rank}(\mathbf{WX}) + \mathcal{R}, \\
		&\text{s.t.} \ \|\mathbf{WX} - \mathbf{D}\|_F \leq \delta_1,
		\|\ \hat{\mathbf{D}} - \mathbf{D} + \mathbf{E}\| \leq \delta_2,
	\end{aligned}
\end{equation}
where  \(\mathbf{D}\in \mathbb{R}^{m \times n}\) is the recovered label distribution,   \(\mathbf{E} \in \mathbb{R}^{m \times n}\) is the annotation bias, and \(\mathcal{R}\) applies regularization. \(\delta_1\) and \(\delta_2\) represent the reconstruction errors in the optimization process.

\begin{figure}[!t]
	\centering
	\includegraphics[width=1\linewidth]{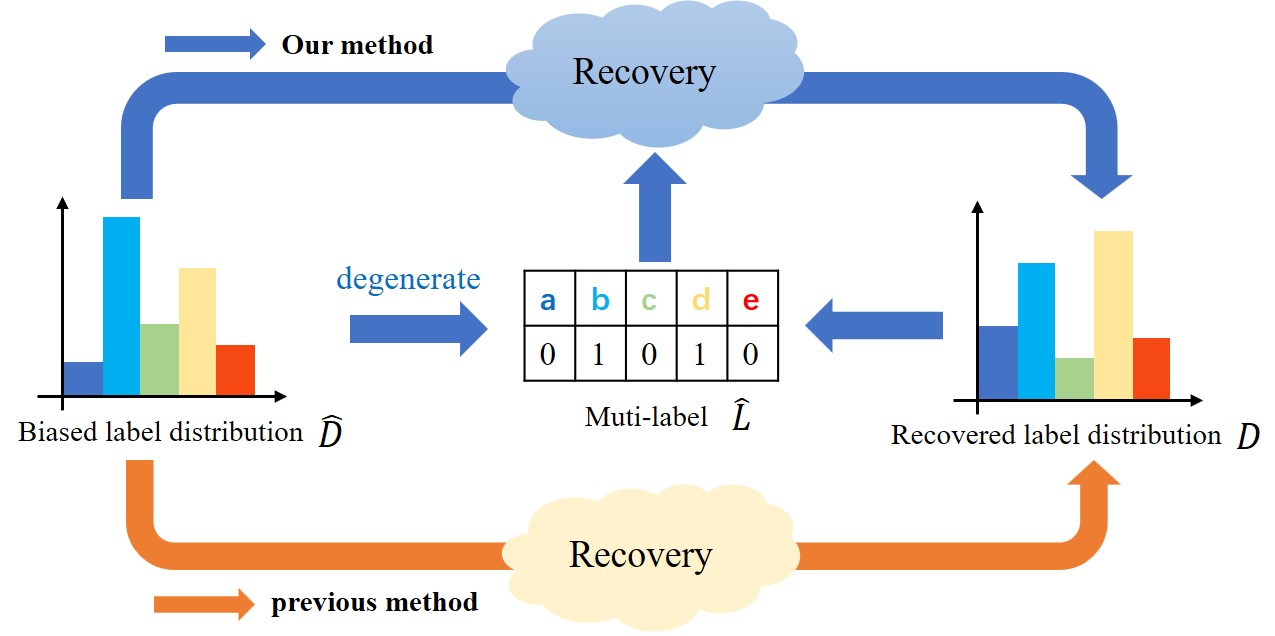}
	\caption{An overview of the proposed BLDL framework. }
	\label{framework}
\end{figure}

However, recent evidence shows that the label distribution tends to be \textit{full-rank} \cite{wang2021label}, and naive apply of low-rank approximation on biased observation leads to \textit{inaccurate recovery and performance degradation}.   In this paper, we address the above limitations of existing methods. Our method are enhance as  below:
	\begin{equation}
		\begin{aligned}
			&\min_{\mathbf{W}, \mathbf{O}, \mathbf{D}}  \text{rank}(\mathbf{W}\mathbf{X}\mathbf{O})+ \alpha\|\hat{\mathbf{D}}\mathbf{O}-\hat{\mathbf{L}}\|_F 	\\
			& + \beta\|\mathbf{W}\mathbf{X} - \mathbf{D}\|_F
			+ \lambda_1\|\mathbf{W}\|_F^2
			+ \lambda_2\|\mathbf{O}\|_F^2,\\
			&\text{s.t.}
			\|\hat{\mathbf{D}} - \mathbf{D}\|_F \leq \delta_1, \quad  \|\mathbf{D}\mathbf{O} - \hat{\mathbf{L}}\|_F \leq \delta_2,
		\end{aligned}
		\label{finallosss}
\end{equation}
where \(\mathbf{O} \in \mathbb{R}^{m \times m}\) models the degradation from label distribution to multi-label, and \(\hat{\mathbf{L}}\footnote{The generation of \(\hat{\mathbf{L}}\) is  described in the Appendix.} \in \mathbb{R}^{m \times n}\) represents the multi-label derived from biased distribution. \(\mathbf{D}\) is recovered label distribution. Parameters \(\alpha\), \(\beta\), and \(\lambda_i\) control the term weightings.  Our method consists of three parts. The \textit{first part} is label distribution recovery, where we use both the biased label distribution and its multi-label representation. Recovering the label distribution via the multi-label space is more reliable than relying directly on the biased distribution, as the discrepancy between the biased and true distributions is much larger than between their multi-label representations. By constraining the degradation of the recovered true distribution to the multi-label space (Condition 2), the reliability of the process is ensured.  We also  require that the difference between the recovered true distribution and the biased distribution be bounded by \(\delta_1\) (Condition 1) to control the recovery of the label distribution.  The \textit{second part} is label distribution learning, where we impose a low-rank constraint on the multi-label space (instead of the label distribution space, since it is full-rank) to capture label correlations (the first, third, and fourth terms of Eq.~(\ref{finallosss})). The \textit{final part} is the multi-label mapping process, where we learn the mapping from multi-labels to label distributions (the second and fifth terms of Eq.~(\ref{finallosss})). The algorithm  flow chart is shown in Fig. \ref{framework}, 

\section{Optimization}

To solve model~(\ref{finallosss}), we relax the rank by its convex alternative, nuclear norm \cite{gu2014weighted}, and then apply the ADMM \cite{8186925} for efficient optimization. The corresponding augmented Lagrangian function is:

	\begin{equation*}
		\begin{aligned}
			&\mathcal{L}(\mathbf{W}, \mathbf{O}, \mathbf{D}, \mathbf{Z}, \mathbf{\Lambda})\\
			&=
			\|\mathbf{Z}\|_*
			+ \alpha\|\mathbf{W}\mathbf{X} - \mathbf{D}\|_F^2
			+ \beta\|\hat{\mathbf{D}}\mathbf{O} - \hat{\mathbf{L}}\|_F^2 \\
			& + \gamma\|\mathbf{D}\mathbf{O} - \hat{\mathbf{L}}\|_F^2
			+ \eta\|\mathbf{D} - \hat{\mathbf{D}}\|_F^2 + \lambda_1\|\mathbf{W}\|_F^2
			+ \lambda_2\|\mathbf{O}\|_F^2 \\
			& + \langle \mathbf{\Lambda}, \mathbf{Z} - \mathbf{W}\mathbf{X}\mathbf{O}\rangle +\frac{\rho}{2}\|\mathbf{Z} - \mathbf{W}\mathbf{X}\mathbf{O}\|_F^2,
		\end{aligned}
	\end{equation*}
where $\mathbf{Z}$ is a splitting variable of \(\mathbf{W}\mathbf{X}\mathbf{O}\), \( \mathbf{\Lambda} \) is  the Lagrange multiplier, and \( \rho \) is positive penalty parameter. The optimization is performed by iteratively updating \( \mathbf{W}, \mathbf{O}, \mathbf{D}, \mathbf{Z}, \boldsymbol{\Lambda} \) as follows:

1) \textbf{W-subproblem} is formulated as:

	\begin{equation*}
		\begin{aligned}
			\mathbf{W}^{k+1} = &\arg\min_{\mathbf{W}} \alpha\|\mathbf{W}\mathbf{X} - \mathbf{D}\|_F^2 + \lambda_1\|\mathbf{W}\|_F^2 \\
			&+ \frac{\rho}{2}\|\mathbf{Z} - \mathbf{W}\mathbf{X}\mathbf{O} + {\boldsymbol{\Lambda}}/{\rho}\|_F^2.
		\end{aligned}
	\end{equation*}

To obtain the  solution, we set the derivative of the objective with respect to \(\mathbf{W}\) to zero, and we get:

	\begin{equation*}
		\begin{aligned}
			\mathbf{W}^{k+1} = &\left(2\alpha \mathbf{D}\mathbf{X}^\top + \rho (\mathbf{Z} + {\boldsymbol{\Lambda}}/{\rho})\mathbf{O}^\top\mathbf{X}^\top\right)\\
			&\cdot \left(2\alpha \mathbf{X}\mathbf{X}^\top + \rho \mathbf{X}\mathbf{O}\mathbf{O}^\top\mathbf{X}^\top + 2\lambda_1 \mathbf{I}\right)^{-1}.
		\end{aligned}
	\end{equation*}

2) \textbf{O-subproblem} is formulated as:

	\begin{equation*}
		\begin{aligned}
			\mathbf{O}^{k+1} =& \arg\min_{\mathbf{O}} \beta\|\hat{\mathbf{D}}\mathbf{O} - \hat{\mathbf{L}}\|_F^2 + \gamma\|\mathbf{D}\mathbf{O} - \hat{\mathbf{L}}\|_F^2 \\
			&+ \lambda_2\|\mathbf{O}\|_F^2 + \frac{\rho}{2}\|\mathbf{Z} - \mathbf{W}\mathbf{X}\mathbf{O} + {\boldsymbol{\Lambda}}/{\rho}\|_F^2.
		\end{aligned}
	\end{equation*}

Similar to the solution for $\mathbf{W}$, and we get:

	\begin{equation*}
		\begin{aligned}
			\mathbf{O}^{k+1} &= \left(2\beta \hat{\mathbf{D}}^\top \hat{\mathbf{D}} + 2\gamma \mathbf{D}^\top \mathbf{D} + \rho \mathbf{X}^\top\mathbf{W}^\top \mathbf{W}\mathbf{X} + 2\lambda_2 \mathbf{I}\right)^{-1}\\
			& \cdot \left(2\beta \hat{\mathbf{D}}^\top \hat{\mathbf{L}} + 2\gamma \mathbf{D}^\top \hat{\mathbf{L}} + \rho \mathbf{X}^\top\mathbf{W}^\top (\mathbf{Z} + {\boldsymbol{\Lambda}}/{\rho})\right).
		\end{aligned}
	\end{equation*}

3) \textbf{D-subproblem} is formulated as:

	\begin{equation*}
		\begin{aligned}
			\mathbf{D}^{k+1}
			&= \arg\min_{\mathbf{D}} \alpha\|\mathbf{W}\mathbf{X} - \mathbf{D}\|_F^2 \\
			& \; \; + \gamma\|\mathbf{D}\mathbf{O} - \hat{\mathbf{L}}\|_F^2
			+ \eta\|\mathbf{D} - \hat{\mathbf{D}}\|_F^2.
		\end{aligned}
	\end{equation*}

Similar to the solution above, and we get:

	\begin{equation*}
		\begin{aligned}
			\mathbf{D}^{k+1} = &\left(2\alpha \mathbf{W}\mathbf{X} + 2\gamma \hat{\mathbf{L}}\mathbf{O}^\top + 2\eta \hat{\mathbf{D}}\right)\\
			&\cdot \left(2\alpha \mathbf{I} + 2\gamma \mathbf{O}\mathbf{O}^\top + 2\eta \mathbf{I}\right)^{-1}.
		\end{aligned}
	\end{equation*}

4) \textbf{Z-subproblem} is formulated as:

	\begin{equation*}
		\mathbf{Z}^{k+1} = \arg\min_{\mathbf{Z}} \|\mathbf{Z}\|_* + \frac{\rho}{2}\|\mathbf{Z} - \mathbf{W}\mathbf{X}\mathbf{O} + {\boldsymbol{\Lambda}}/{\rho}\|_F^2.
	\end{equation*}

The solution has closed-form via the soft-thresholding operator \cite{6226423}:

	\begin{equation*}
		\mathbf{Z}^{k+1} = \text{SVT}_{{1}/{\rho}}(\mathbf{W}\mathbf{X}\mathbf{O} - {\boldsymbol{\Lambda}}/{\rho}).
	\end{equation*}

5) Finally, update the \textbf{Lagrange Multipliers}
\begin{equation*}
	\boldsymbol{\Lambda}^{k+1} = \boldsymbol{\Lambda}^k + \rho (\mathbf{Z} - \mathbf{W}\mathbf{X}\mathbf{O}),
\end{equation*}
and update $\rho$ by $\mu\rho$ for some $\mu>1$.

\section{Theoretical Analysis}

We first analyze the convergence \footnote{The proof detail can be found in the appendix.} of the above algorithm in solving the proposed BLDL model.

\begin{theorem}
	All these iterative solutions $\mathbf{W}, \mathbf{O},\mathbf{D},\mathbf{Z}, \mathbf{\Lambda}$ generated by the above ADMM procedure are bounded and convergent.
\end{theorem}

We then establish the generalization error bound for the proposed BLDL framework.

\begin{theorem}
	The generalization error of the model, defined as $\mathcal{E}_{\text{gen}} = \mathbb{E}_{\mathcal{D}}\left[\|\mathbf{W}\mathbf{X} - \mathbf{D}_{\text{true}}\|_F^2\right]$, which is bounded as follows:
	
	\begin{equation*}	\mathcal{E}_{\text{gen}} \leq \frac{\delta_4^2}{1 - \delta} + (\delta_3 + \epsilon)^2 + \mathcal{O}\left(\frac{\text{rank}(\mathbf{W}\mathbf{X}\mathbf{O})}{\sqrt{n}}\right),
	\end{equation*}
	where \(\delta_4\) is the upper bound imposed by the optimization constraint, \(\delta\) is the Restricted Isometry Property (RIP) constant, \(\delta_3\) bounds the deviation between \(\mathbf{D}\) and \(\hat{\mathbf{D}}\), and \(\mathcal{O}\left(\frac{\text{rank}(\mathbf{W}\mathbf{X}\mathbf{O})}{\sqrt{n}}\right)\) accounts for the complexity induced by nuclear norm regularization.
\end{theorem}

According to this theorem, it ensures that the recovered label distribution is robust to noise, consistent with biased observations, and achieves model simplicity via low-rank constraints.

\section{Experiments}
\textbf{5.1 Experimental Configuration}

\textbf{Experimental Datasets:} We evaluate our proposed method on 12 real-world datasets with label distributions. The details are summarized in the Appendix.

\textbf{Comparing Methods:} We compare our proposed method with seven state-of-the-art LDL methods, briefly introduced as follows:
\textit{DN-ILDL} \cite{kou2024inaccurate}: Handles label-dependent and instance-dependent noise by utilizing linear mappings, group sparsity, and graph regularization.
\textit{LDL-SCL} \cite{zheng2018label}: Explores local sample correlations through the construction of a local correlation vector.
\textit{LDLLC} \cite{jia2018label}: Captures label correlations via a distance-based mapping function.
\textit{LRS-LDL} \cite{kou2023instance}: Learns a low-rank linear mapping for ground truth and a sparse mapping for noise.
\textit{LDLLDM} \cite{wang2021label}: Models both global and local label correlations by learning the underlying manifold structure of label distributions.
\textit{EDL-LRL} \cite{jia2019facial}: Utilizes local label correlations to effectively capture varying intensities of multiple emotions.
\textit{TLRLDL} \cite{ijcai2024p478}: Integrates an auxiliary multi-label learning (MLL) process within LDL to capture low-rank label correlations.

\textbf{5.2 Results and Discussion}
Parameter \(C\) was set to 0.1, 0.2, and 0.3 to generate biased label distributions. Table \ref{zhushiyan} shows the experimental results (mean\(\pm\)std) of various methods on ID.1-8 datasets for Clark and Cosine metrics\footnote{Full results for all datasets and metrics are in the appendix.}. For \(C=0.1\), the Friedman test \cite{demvsar2006statistical} rejected the null hypothesis that \textit{all methods perform equally} (Table \ref{criticalFF}). Subsequently, the Bonferroni-Dunn test \cite{demvsar2006statistical} was used to compare BLDL with others, where methods differing by more than one Critical Difference (CD) are considered significantly different. CD diagrams in Fig. \ref{CD} highlight methods within one CD of BLDL connected by a thin line, confirming BLDL's significant advantage. From these results, we conclude:

\textbf{(i)} BLDL achieves Top-1 performance in \textit{85.42\% (41/48)} of configurations and consistently ranks first across all metrics by effectively addressing \textit{bias} in the learning process.

\textbf{(ii)} Compared to methods focusing solely on label correlations (e.g., LDLLC, LDLLDM), BLDL performs better as these methods ignore \textit{bias} in label distributions.

\textbf{(iii)} BLDL outperforms methods addressing \textit{bias} alone (e.g., DN-ILDL, LRS-LDL) by leveraging label correlations, which these methods fail to exploit effectively.

\begin{figure}[!t]
	\centering
	\begin{minipage}{0.25\textwidth}
		\centering
		\includegraphics[width=\textwidth]{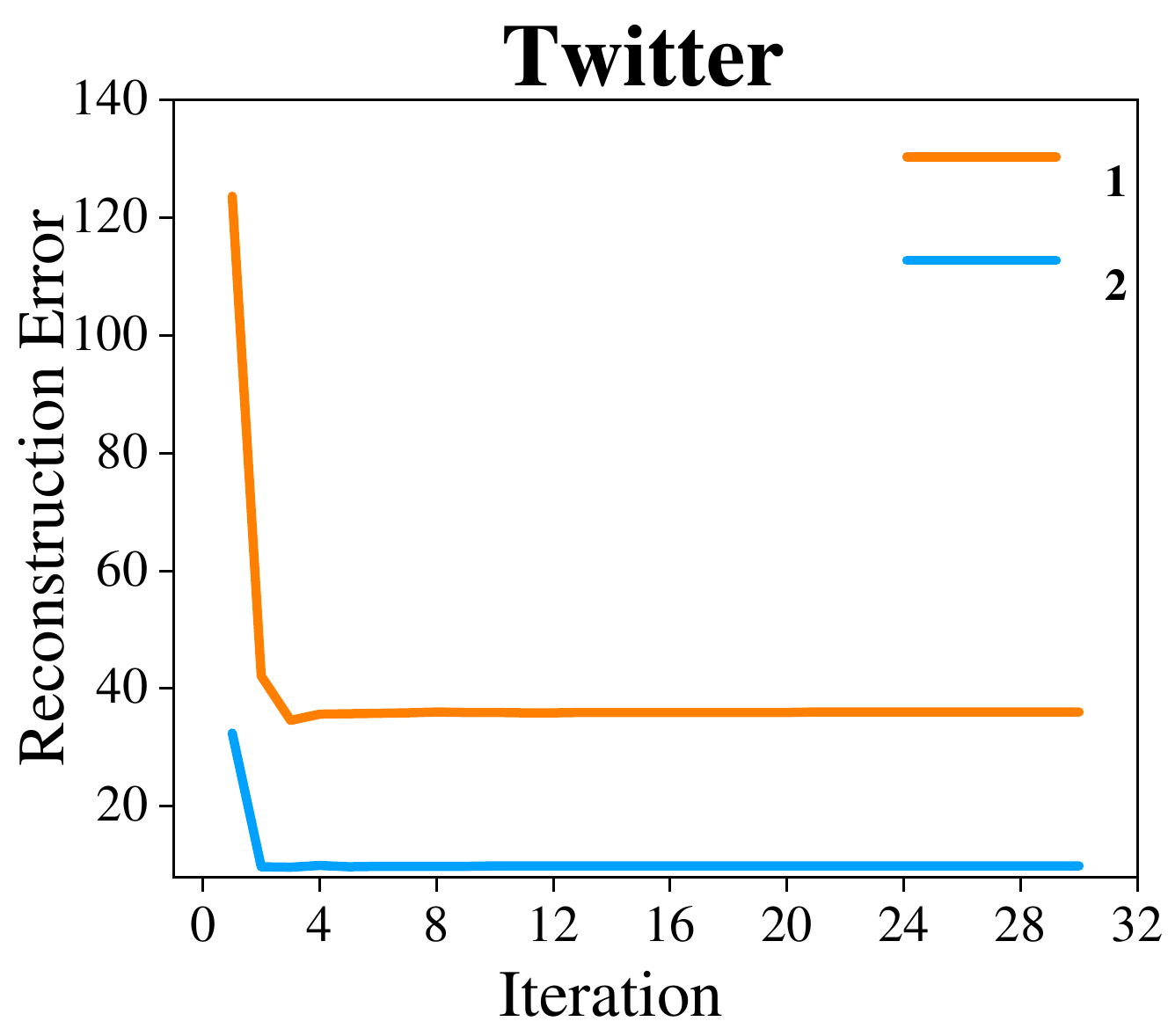}
		
	\end{minipage}%
	\begin{minipage}{0.25\textwidth}
		\centering
		\includegraphics[width=\textwidth]{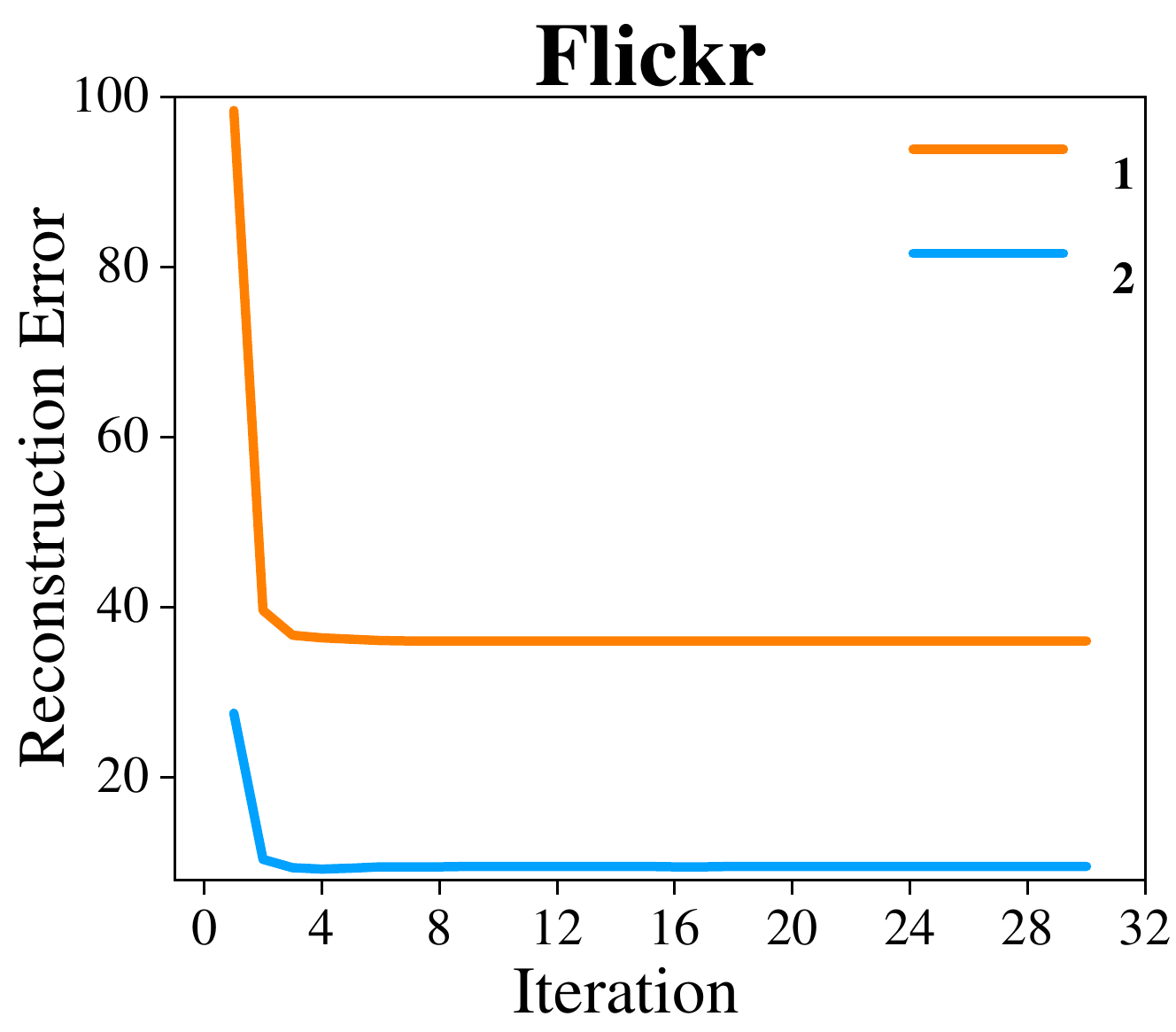}
		
	\end{minipage}
	\caption{Error reduction ($\delta_1$ and $\delta_2$) during iterations on the Twitter and Flickr datasets. }
	
	\label{jiashe}
\end{figure}

\begin{figure}[!t]
	\centering
	\begin{minipage}{0.25\textwidth}
		\centering
		\includegraphics[width=\textwidth]{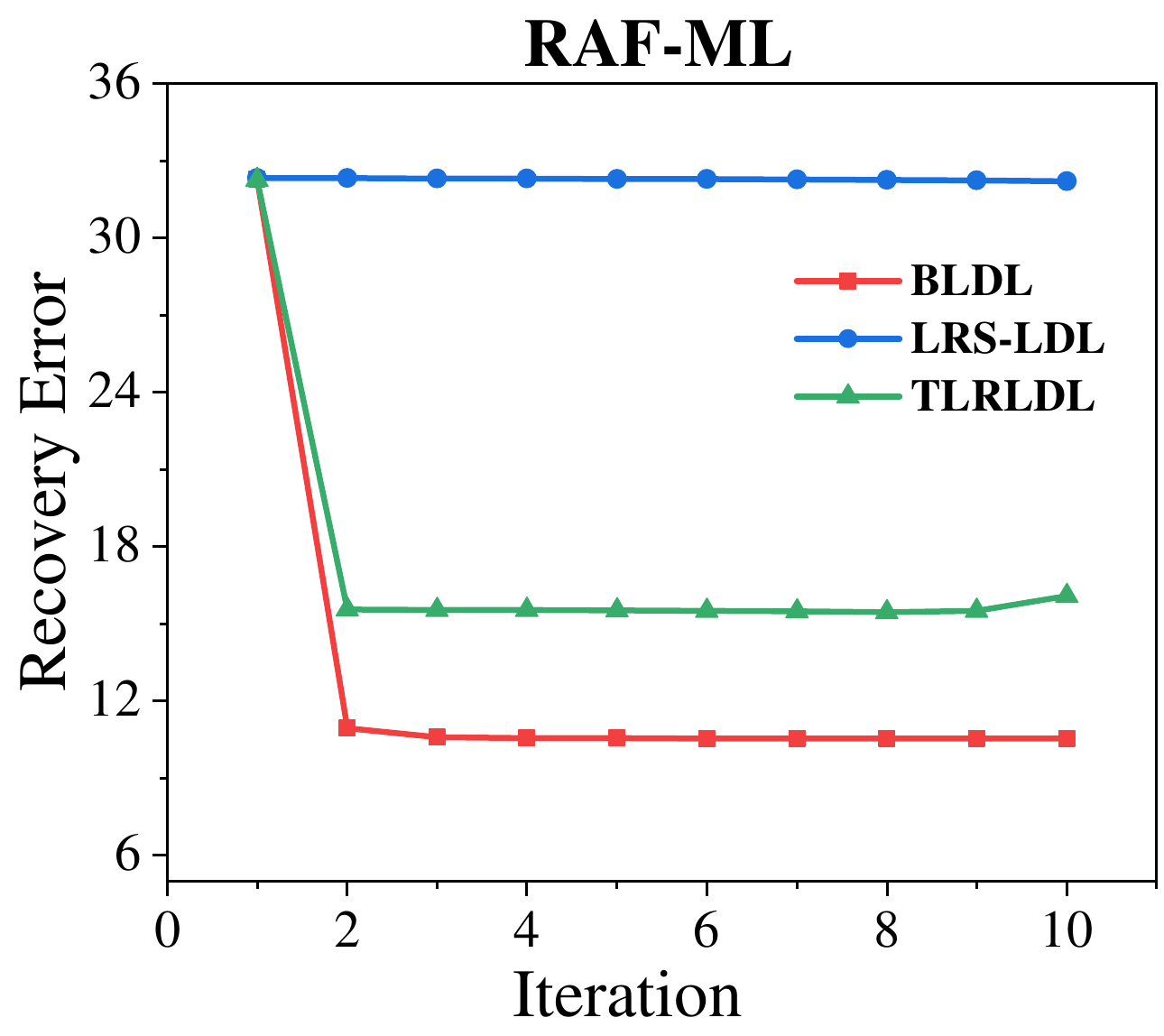}
		
	\end{minipage}%
	\begin{minipage}{0.25\textwidth}
		\centering
		\includegraphics[width=\textwidth]{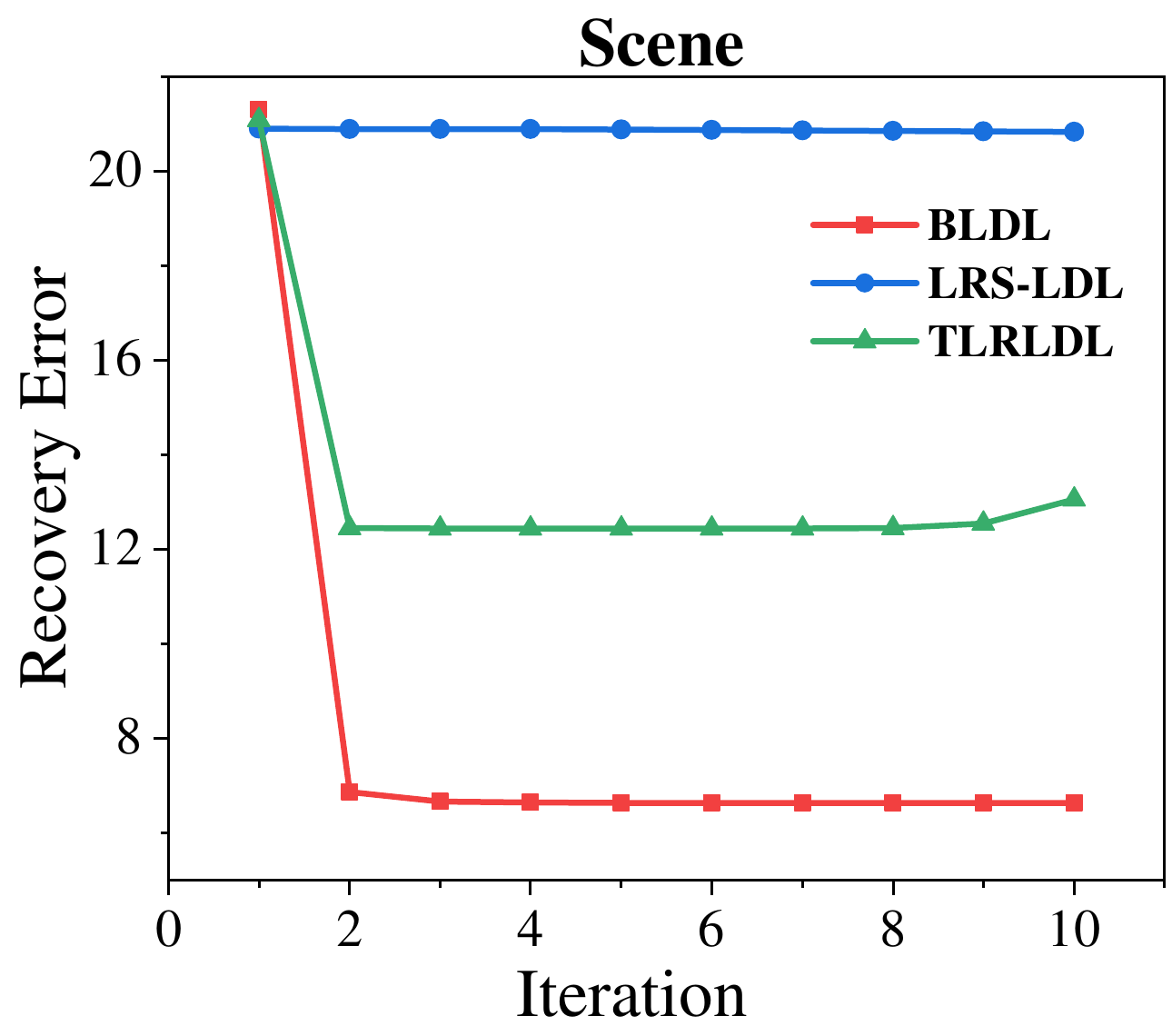}
		
	\end{minipage}
	\caption{Reconstruction error in recovering true distributions for different methods during the traning stage.}
	\label{huifu}
\end{figure}

\textbf{5.3 Ablation Study}:
To validate the proposed method, we design two ablated versions:
\textbf{(i)} \textit{BLDL-a}: Removes the biased label distribution recovery process:

	\begin{equation*}
		\begin{aligned}
			\min_{\mathbf{W},\mathbf{O}}
			&\frac{1}{2}\alpha \|\mathbf{WX} - \hat{\mathbf{D}}\|_F^2 + \frac{1}{2}\beta \|\hat{\mathbf{D}}\mathbf{O} - \hat{\mathbf{L}}\|_F^2 \\
			&+ \|\mathbf{WXO}\|_* + \lambda_1\|\mathbf{W}\|_F^2 + \lambda_2\|\mathbf{O}\|_F^2.
		\end{aligned}
	\end{equation*}

\textbf{(ii)} \textit{BLDL-b}: Replaces the low-rank constraint on \(\mathbf{WXO}\) with one on \(\mathbf{WO}\), applying low-rank modeling in the label distribution space instead of the multi-label space.

We compare BLDL-a and BLDL-b with BLDL using biased label distributions generated with \(C=0.1\). Figure~\ref{xiaorongkeshihua} shows results on seven datasets for Chebyshev and Intersection metrics. Wilcoxon signed-rank tests \cite{demvsar2006statistical} confirm the statistical significance of BLDL over BLDL-a and BLDL-b (Table~\ref{w-test}). According to Figure \ref{xiaorongkeshihua} and Table \ref{w-test}, the following conclusions are drawn:

\textbf{(i)} \textit{BLDL-a vs. BLDL}: Removing the bias recovery process significantly degrades performance, highlighting the necessity of explicitly modeling the \textit{bias} for accurate label recovery.

\textbf{(ii)} \textit{BLDL-b vs. BLDL}: Using low-rank constraints on $\mathbf{WXO}$ improves results, as the multi-label space is low-rank and more robust to noise and bias.

\textbf{(iii)} \textit{Overall}: BLDL consistently outperforms both ablated versions, confirming the importance of addressing \textit{bias} and low-rank \textit{label correlation}.

\textbf{Validation of Hypothesis}: We computed the difference between the recovered label distribution and the biased label distribution during the training phase, denoted as \( \delta_1 \), and the difference between the multi-label corresponding to the recovered label distribution and the multi-label corresponding to the biased distribution, denoted as \( \delta_2 \). As show in Fig. \ref{jiashe}, it can be observed that \( \delta_1 \) is consistently greater than \( \delta_2 \) until convergence. Therefore, during the label distribution recovery phase, multi-label information is more reliable than the biased label distribution information.

\textbf{Analysis of Label Distribution Recovery:}  
We evaluate the recovery error of label distributions during training using the Frobenius norm \( \| D_{\text{recover}} - D_{\text{truth}} \|_F \), as shown in Fig.~\ref{huifu}. Our method achieves lower recovery error compared to LRS-LDL, demonstrating the benefit of leveraging multi-label information. Additionally, it outperforms TLRLDL by addressing biases in the learning process, further improving recovery accuracy.

\textbf{5.4 Parameter Sensitivity Analysis}:
We analyze the sensitivity of \(\alpha, \beta, \eta, \lambda_1\), with \(\alpha, \beta, \lambda_1\) chosen from \{0.1, 0.05, 0.01, 0.005, 0.001\} and \(\eta\) from \{1, 10, 50, 100, 150\}. Experiments on five datasets (Scene, SBU, Emotion, Spo, Spoem) show in Fig. \ref{canshufenxi} that BLDL exhibits stable performance, demonstrating robustness to parameter variations.

\begin{table*}
	\centering
	\scriptsize
	\renewcommand{\arraystretch}{1.2}
	\setlength{\tabcolsep}{1.05mm}
	\begin{tabular}{@{}l|c|l|cccccccc@{}}
		\toprule
		& C & Metric & BLDL & DN-ILDL & LDL-SCL & LDLLC & LRS-LDL & LDLDM & EDL-LDL & TLRLDL \\
		\midrule
		\multirow{6}{*}{Fli} & \multirow{2}{*}{0.1} & Clark & \textbf{2.1146$\pm$.0020} & 2.1990$\pm$.0008 & 2.1576$\pm$.0001 & 2.1990$\pm$.0042 & 2.2029$\pm$.0028 & 2.1705$\pm$.0001 & 2.1797$\pm$.0001 & 2.1777$\pm$.0005 \\
		& & Cosine & \textbf{0.8368$\pm$.0003} & 0.5516$\pm$.0002 & 0.7356$\pm$.0001 & 0.5781$\pm$.0038 & 0.5537$\pm$.0005 & 0.6964$\pm$.0001 & 0.7436$\pm$.0001 & 0.6421$\pm$.0009 \\
		\cmidrule{2-11}
		& \multirow{2}{*}{0.2}
		& Clark & \textbf{2.1175$\pm$.0010} & 2.1969$\pm$.0153 & 2.1587$\pm$.0001 & 2.1984$\pm$.0021 & 2.1984$\pm$.0078 & 2.1707$\pm$.0001 & 2.1758$\pm$.0001 & 2.1768$\pm$.0020\\
		& & Cosine & \textbf{0.8316$\pm$.0034} & 0.5531$\pm$.0052 & 0.7289$\pm$.0001 & 0.5781$\pm$.0004 & 0.5556$\pm$.0027 & 0.6869$\pm$.0001 & 0.7414$\pm$.0001 & 0.6356$\pm$.0006\\
		\cmidrule{2-11}
		& \multirow{2}{*}{0.3}
		& Clark & \textbf{2.1382$\pm$.0010} & 2.2018$\pm$.0033 & 2.1597$\pm$.0001 & 2.1990$\pm$.0019 & 2.1985$\pm$.0028 & 2.1713$\pm$.0001 & 2.1733$\pm$.0001 & 2.1778$\pm$.0005\\
		& & Cosine & \textbf{0.8276$\pm$.0021} & 0.5510$\pm$.0017 & 0.7228$\pm$.0001 & 0.5761$\pm$.0008 & 0.5554$\pm$.0001 & 0.6792$\pm$.0001 & 0.7394$\pm$.0001 & 0.6322$\pm$.0014\\
		\midrule
		\multirow{6}{*}{Twi} & \multirow{2}{*}{0.1} & Clark & \textbf{2.2517$\pm$.0028} & 2.4072$\pm$.0037 & 2.3602$\pm$.0001 & 2.3957$\pm$.0039 & 2.4043$\pm$.0040 & 2.3656$\pm$.0001 & 2.3874$\pm$.0001 & 2.3726$\pm$.0001 \\
		& & Cosine & \textbf{0.8582$\pm$.0025} & 0.4955$\pm$.0013 & 0.7537$\pm$.0001 & 0.5617$\pm$.0022 & 0.5000$\pm$.0008 & 0.7519$\pm$.0001 & 0.8134$\pm$.0001 & 0.6283$\pm$.0004 \\
		\cmidrule{2-11}
		& \multirow{2}{*}{0.2}
		& Clark & \textbf{2.2783$\pm$.0113} & 2.4014$\pm$.0008 & 2.3608$\pm$.0001 & 2.3897$\pm$.0036 & 2.4015$\pm$.0020 & 2.3651$\pm$.0001 & 2.3827$\pm$.0001 & 2.3763$\pm$.0069\\
		& & Cosine & \textbf{0.8535$\pm$.0015} & 0.4980$\pm$.0009 & 0.7509$\pm$.0001 & 0.5621$\pm$.0043 & 0.5016$\pm$.0017 & 0.7372$\pm$.0001 & 0.8124$\pm$.0001 & 0.6211$\pm$.0037\\
		\cmidrule{2-11}
		& \multirow{2}{*}{0.3}
		& Clark & \textbf{2.3051$\pm$.0080} & 2.4043$\pm$.0050 & 2.3616$\pm$.0001 & 2.3906$\pm$.0034 & 2.4016$\pm$.0020 & 2.3657$\pm$.0001 & 2.3790$\pm$.0001 & 2.3780$\pm$.0017\\
		& & Cosine & \textbf{0.8480$\pm$.0012} & 0.4957$\pm$.0018 & 0.7447$\pm$.0001 & 0.5575$\pm$.0029 & 0.5014$\pm$.0016 & 0.7274$\pm$.0001 & 0.8106$\pm$.0001 & 0.6131$\pm$.0003\\
		\midrule
		\multirow{6}{*}{Emo} & \multirow{2}{*}{0.1} & Clark & \textbf{1.6237$\pm$.0067} & 1.6753$\pm$.0032 & 1.6457$\pm$.0001 & 1.6875$\pm$.0077 & 1.6820$\pm$.0113 & 1.6404$\pm$.0002 & 1.6423$\pm$.0001 & 1.6697$\pm$.0114 \\
		& & Cosine & 0.7462$\pm$.0099 & 0.6592$\pm$.0008 & 0.7266$\pm$.0001 & 0.6700$\pm$.0005 & 0.6560$\pm$.0047 & 0.7417$\pm$.0001 & \textbf{0.7639$\pm$.0001} & 0.6841$\pm$.0020 \\
		\cmidrule{2-11}
		& \multirow{2}{*}{0.2}
		& Clark & \textbf{1.6408$\pm$.0008} & 1.6716$\pm$.0174 & 1.6480$\pm$.0001 & 1.6840$\pm$.0106 & 1.6774$\pm$.0031 & 1.6432$\pm$.0002 & 1.6412$\pm$.0002 & 1.6677$\pm$.0020\\
		& & Cosine & 0.7499$\pm$.0004 & 0.6613$\pm$.0042 & 0.7227$\pm$.0001 & 0.6701$\pm$.0063 & 0.6565$\pm$.0005 & 0.7370$\pm$.0001 & \textbf{0.7589$\pm$.0001} & 0.6816$\pm$.0009\\
		\cmidrule{2-11}
		& \multirow{2}{*}{0.3}
		& Clark & \textbf{1.6295$\pm$.0004} & 1.6753$\pm$.0051 & 1.6491$\pm$.0001 & 1.6848$\pm$.0104 & 1.6775$\pm$.0031 & 1.6467$\pm$.0001 & 1.6402$\pm$.0002 & 1.6799$\pm$.0001\\
		& & Cosine & 0.7458$\pm$.0030 & 0.6579$\pm$.0004 & 0.7193$\pm$.0001 & 0.6685$\pm$.0059 & 0.6564$\pm$.0005 & 0.7304$\pm$.0001 & \textbf{0.7574$\pm$.0001} & 0.6771$\pm$.0014\\
		\midrule
		\multirow{6}{*}{Fbp} & \multirow{2}{*}{0.1} & Clark & \textbf{1.2494$\pm$.0047} & 1.5052$\pm$.0013 & 1.4032$\pm$.0001 & 1.4866$\pm$.0010 & 1.5039$\pm$.0041 & 1.4391$\pm$.0001 & 1.3360$\pm$.0001 & 1.6697$\pm$.0021 \\
		& & Cosine & \textbf{0.9342$\pm$.0010} & 0.6571$\pm$.0005 & 0.8596$\pm$.0001 & 0.6772$\pm$.0035 & 0.6617$\pm$.0014 & 0.7929$\pm$.0001 & 0.9169$\pm$.0001 & 0.6841$\pm$.0004 \\
		\cmidrule{2-11}
		& \multirow{2}{*}{0.2}
		& Clark & \textbf{1.2974$\pm$.0024} & 1.5054$\pm$.0021 & 1.4115$\pm$.0001 & 1.4891$\pm$.0001 & 1.5027$\pm$.0003 & 1.4475$\pm$.0001 & 1.3472$\pm$.0001 & 1.4579$\pm$.0015\\
		& & Cosine & \textbf{0.9264$\pm$.0012} & 0.6571$\pm$.0003 & 0.8478$\pm$.0001 & 0.6754$\pm$.0026 & 0.6625$\pm$.0009 & 0.7795$\pm$.0001 & 0.9127$\pm$.0001 & 0.7619$\pm$.0027\\
		\cmidrule{2-11}
		& \multirow{2}{*}{0.3}
		& Clark & \textbf{1.3084$\pm$.0053} & 1.5040$\pm$.0009 & 1.4182$\pm$.0001 & 1.4895$\pm$.0003 & 1.5028$\pm$.0003 & 1.4530$\pm$.0001 & 1.3542$\pm$.0001 & 1.4623$\pm$.0004\\
		& & Cosine & \textbf{0.9251$\pm$.0007} & 0.6577$\pm$.0004 & 0.8367$\pm$.0001 & 0.6747$\pm$.0022 & 0.6622$\pm$.0009 & 0.7695$\pm$.0001 & 0.9083$\pm$.0001 & 0.7553$\pm$.0005\\
		\midrule
		\multirow{6}{*}{Scu} & \multirow{2}{*}{0.1} & Clark & \textbf{1.3869$\pm$.0051} & 1.4955$\pm$.0044 & 1.4494$\pm$.0001 & 1.4838$\pm$.0056 & 1.4985$\pm$.0031 & 1.4265$\pm$.0001 & 1.3908$\pm$.0001 & 1.4735$\pm$.0036 \\
		& & Cosine & \textbf{0.8409$\pm$.0075} & 0.6647$\pm$.0016 & 0.7775$\pm$.0001 & 0.6717$\pm$.0068 & 0.6672$\pm$.0008 & 0.8089$\pm$.0001 & 0.8405$\pm$.0001 & 0.7217$\pm$.0028 \\
		\cmidrule{2-11}
		& \multirow{2}{*}{0.2}
		& Clark & \textbf{1.3936$\pm$.0090} & 1.4930$\pm$.0037 & 1.4540$\pm$.0001 & 1.4796$\pm$.0055 & 1.4999$\pm$.0003 & 1.4357$\pm$.0001 & 1.3965$\pm$.0001 & 1.4723$\pm$.0045\\
		& & Cosine & 0.8344$\pm$.0013 & 0.6653$\pm$.0013 & 0.7761$\pm$.0001 & 0.6687$\pm$.0056 & 0.6665$\pm$.0018 & 0.7965$\pm$.0001 & \textbf{0.8369$\pm$.0001} & 0.7213$\pm$.0010\\
		\cmidrule{2-11}
		& \multirow{2}{*}{0.3}
		& Clark & \textbf{1.3934$\pm$.0066} & 1.4971$\pm$.0049 & 1.4562$\pm$.0001 & 1.4807$\pm$.0048 & 1.5001$\pm$.0005 & 1.4394$\pm$.0001 & 1.4022$\pm$.0001 & 1.4748$\pm$.0005\\
		& & Cosine & \textbf{0.8341$\pm$.0032} & 0.6634$\pm$.0013 & 0.7683$\pm$.0001 & 0.6669$\pm$.0069 & 0.6662$\pm$.0019 & 0.7908$\pm$.0001 & 0.834$\pm$.0001 & 0.7131$\pm$.0025\\
		\midrule
		\multirow{6}{*}{Raf} & \multirow{2}{*}{0.1} & Clark & \textbf{1.3864$\pm$.0003} & 1.6119$\pm$.0037 & 1.5534$\pm$.0001 & 1.6115$\pm$.0027 & 1.6102$\pm$.0012 & 1.6025$\pm$.0001 & 1.5410$\pm$.0001 & 1.5688$\pm$.0031 \\
		& & Cosine & \textbf{0.8734$\pm$.0006} & 0.6413$\pm$.0006 & 0.7433$\pm$.0001 & 0.6360$\pm$.0011 & 0.6455$\pm$.0001 & 0.6541$\pm$.0001 & 0.7580$\pm$.0001 & 0.7200$\pm$.0011 \\
		\cmidrule{2-11}
		& \multirow{2}{*}{0.2}
		& Clark & \textbf{1.4205$\pm$.0050} & 1.6082$\pm$.0005 & 1.5586$\pm$.0001 & 1.6071$\pm$.0021 & 1.6043$\pm$.0034 & 1.6015$\pm$.0001 & 1.5464$\pm$.0001 & 1.5723$\pm$.0052\\
		& & Cosine & \textbf{0.8630$\pm$.0007} & 0.6418$\pm$.0006 & 0.7349$\pm$.0001 & 0.6356$\pm$.0031 & 0.6472$\pm$.0018 & 0.6566$\pm$.0001 & 0.7506$\pm$.0001 & 0.7153$\pm$.0016\\
		\cmidrule{2-11}
		& \multirow{2}{*}{0.3}
		& Clark & \textbf{1.4270$\pm$.0019} & 1.6098$\pm$.0005 & 1.5638$\pm$.0001 & 1.6076$\pm$.0019 & 1.6044$\pm$.0033 & 1.6034$\pm$.0001 & 1.5496$\pm$.0001 & 1.5711$\pm$.0001\\
		& & Cosine & \textbf{0.8519$\pm$.0010} & 0.6417$\pm$.0001 & 0.7260$\pm$.0001 & 0.6352$\pm$.0029 & 0.6469$\pm$.0018 & 0.6532$\pm$.0001 & 0.7469$\pm$.0001 & 0.7103$\pm$.0022\\
		\midrule
		\multirow{6}{*}{Gen} & \multirow{2}{*}{0.1} & Clark & \textbf{2.1149$\pm$.0179} & 2.1151$\pm$.0006 & 2.1230$\pm$.0001 & 2.1618$\pm$.0087 & 2.1222$\pm$.0069 & 2.1248$\pm$.0001 & 2.1240$\pm$.0003 & 2.1312$\pm$.0030 \\
		& & Cosine & \textbf{0.8353$\pm$.0023} & 0.8342$\pm$.0003 & 0.8337$\pm$.0001 & 0.8196$\pm$.0006 & 0.8342$\pm$.0009 & 0.8338$\pm$.0001 & 0.8339$\pm$.0001 & 0.8317$\pm$.0005 \\
		\cmidrule{2-11}
		& \multirow{2}{*}{0.2}
		& Clark & \textbf{2.1228$\pm$.0107} & 2.1246$\pm$.0066 & 2.1237$\pm$.0001 & 2.1591$\pm$.0018 & 2.1256$\pm$.0222 & 2.1245$\pm$.0001 & 2.1236$\pm$.0001 & 2.1312$\pm$.0127\\
		& & Cosine & \textbf{0.8340$\pm$.0016} & 0.8336$\pm$.0021 & 0.8333$\pm$.0001 & 0.8202$\pm$.0014 & 0.8334$\pm$.0028 & 0.8337$\pm$.0001 & 0.8338$\pm$.0001 & 0.8329$\pm$.0016\\
		\cmidrule{2-11}
		& \multirow{2}{*}{0.3}
		& Clark & \textbf{2.1225$\pm$.0035} & 2.1269$\pm$.0035 & 2.1225$\pm$.0005 & 2.1591$\pm$.0018  & 2.1256$\pm$.0222 & 2.1249$\pm$.0006 & 2.1240$\pm$.0005 & 2.1296$\pm$.0187\\
		& & Cosine & \textbf{0.8338$\pm$.0003} & 0.8335$\pm$.0007 & 0.8333$\pm$.0001 & 0.8202$\pm$.0014 & 0.8334$\pm$.0028 & 0.8336$\pm$.0001 & 0.8337$\pm$.0001 & 0.8334$\pm$.0012\\
		\midrule
		\multirow{6}{*}{Sce} & \multirow{2}{*}{0.1} & Clark & 2.4900$\pm$.0012 & 2.4848$\pm$.0017 & \textbf{2.4654$\pm$.0001} & 2.4915$\pm$.0019 & 2.4747$\pm$.0012 & 2.4702$\pm$.0001 & 2.4775$\pm$.0001 & 2.4736$\pm$.0001 \\
		& & Cosine & \textbf{0.7037$\pm$.0045} & 0.5748$\pm$.0003 & 0.6770$\pm$.0001 & 0.5572$\pm$.0008 & 0.5806$\pm$.0013 & 0.6431$\pm$.0001 & 0.6500$\pm$.0001 & 0.6253$\pm$.0007 \\
		\cmidrule{2-11}
		& \multirow{2}{*}{0.2}
		& Clark & 2.4808$\pm$.0129 & 2.4829$\pm$.0001 & \textbf{2.4668$\pm$.0001} & 2.4950$\pm$.0066 & 2.4819$\pm$.0059 & 2.4711$\pm$.0001 & 2.4751$\pm$.0001 & 2.4771$\pm$.0040\\
		& & Cosine & \textbf{0.7010$\pm$.0010} & 0.5753$\pm$.0007 & 0.6692$\pm$.0001 & 0.5553$\pm$.0018 & 0.5780$\pm$.0029 & 0.6367$\pm$.0001 & 0.6469$\pm$.0001 & 0.6205$\pm$.0034\\
		\cmidrule{2-11}
		& \multirow{2}{*}{0.3}
		& Clark & 2.5100$\pm$.0123 & 2.4859$\pm$.0057 & \textbf{2.4683$\pm$.0001} & 2.4951$\pm$.0066 & 2.4820$\pm$.0059 & 2.4721$\pm$.0002 & 2.4758$\pm$.0001 & 2.4799$\pm$.0040\\
		& & Cosine & \textbf{0.6876$\pm$.0026} & 0.5738$\pm$.0034 & 0.6611$\pm$.0001 & 0.5551$\pm$.0016 & 0.5779$\pm$.0030 & 0.6328$\pm$.0001 & 0.6455$\pm$.0001 & 0.6168$\pm$.0029\\
		\midrule
		\multicolumn{3}{c|}{top-1 times} & \textbf{41} & 0 & 3 & 0 & 0 & 0 & 4 & 0 \\
		\bottomrule
	\end{tabular}
	\centering
	\caption{Results (mean$\pm$std) of the comparing methods in terms of two metrics on ID.1-8 datasets (each is denoted by its first three letters), where the best results are bolded. }
	\label{zhushiyan}
\end{table*}

\begin{table*}[!h]
	\small
	\centering
	\renewcommand{\arraystretch}{1.2}
	\setlength{\tabcolsep}{2.2mm}
	\begin{tabular}{@{}l|ccccccc@{}}
		\hline
		Critical Value ($\alpha=0.05$) & Evaluation metric & Cheb & Clark & KL & Canber & Intersec & Cosine \\ \hline
		\multicolumn{1}{c|}{2.1310} & Friedman Statistics $F_F$ & 17.6522 & 8.2619 & 12.6744 & 12.5264 & 19.2706 & 22.4616 \\ \hline
	\end{tabular}
	\caption{The Friedman statistics $F_F$
		in terms of six evaluation metrics, as well as the critical value at a significance level of 0.05 (8 algorithms on 12 datasets). }
	\label{criticalFF}
\end{table*}

\begin{figure*}[!h]
	\centering
	\subfloat[Chebyshev]{
		\includegraphics[scale=0.33]{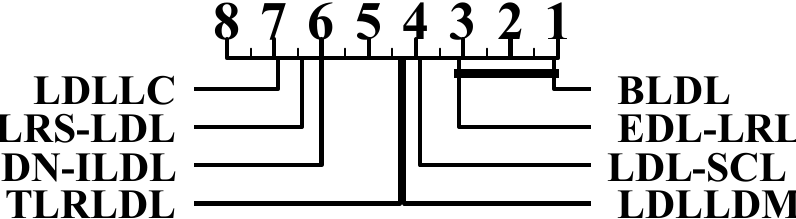}
	}
	\hfil
	\subfloat[Clark]{
		\includegraphics[scale=0.33]{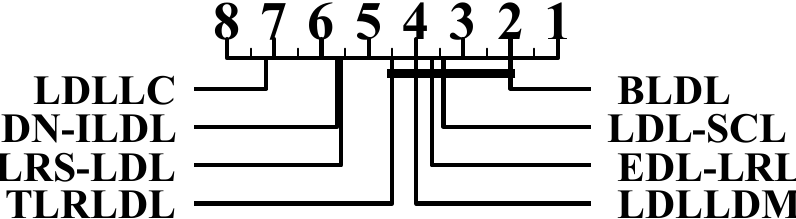}
	}
	\hfil
	\subfloat[KL]{
		\includegraphics[scale=0.33]{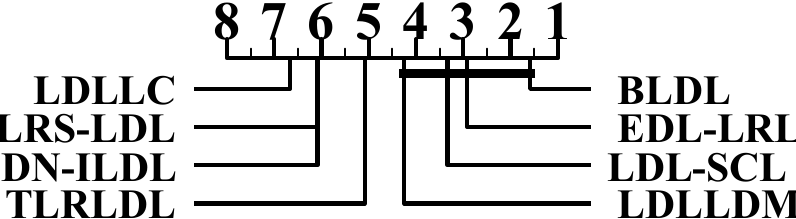}
	}
	\hfil
	\subfloat[Canberra]{
		\includegraphics[scale=0.33]{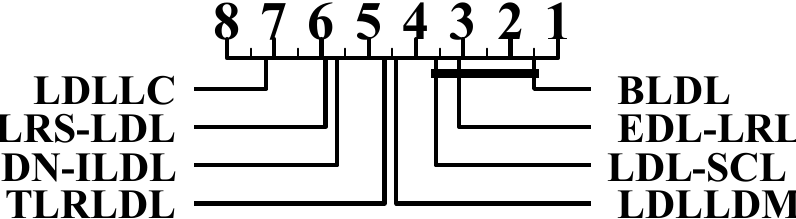}
	}
	\hfil
	\subfloat[Intersection]{
		\includegraphics[scale=0.33]{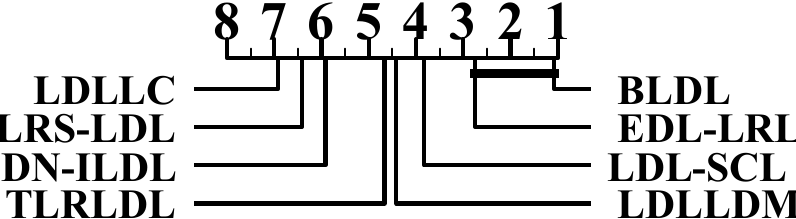}
	}
	\hfil
	\subfloat[Cosine]{
		\includegraphics[scale=0.33]{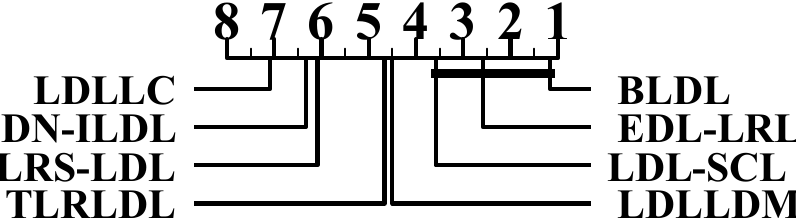}
	}
	\caption{CD diagrams of the comparing methods in terms of each metrics.
		For the tests, CD equals 2.3296 at 0.05 signifcance level.}
	\label{CD}
\end{figure*}

\begin{figure*}[!h]
	\centering
	{
		\includegraphics[scale=0.29]{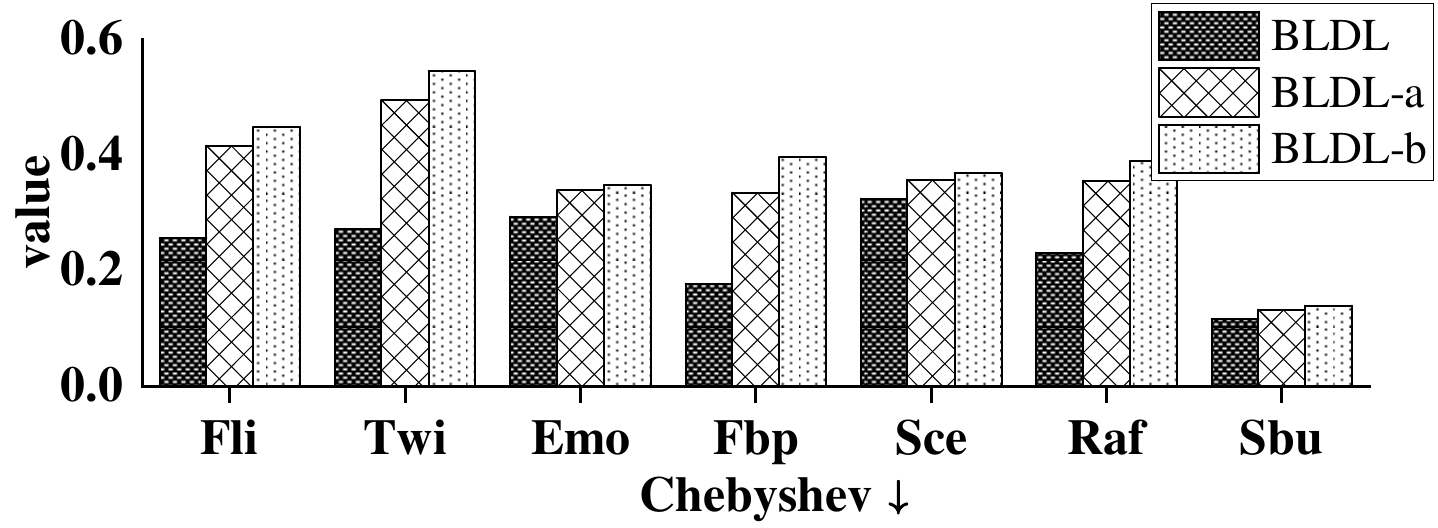}
	}
	\hfil
	{
		\includegraphics[scale=0.29]{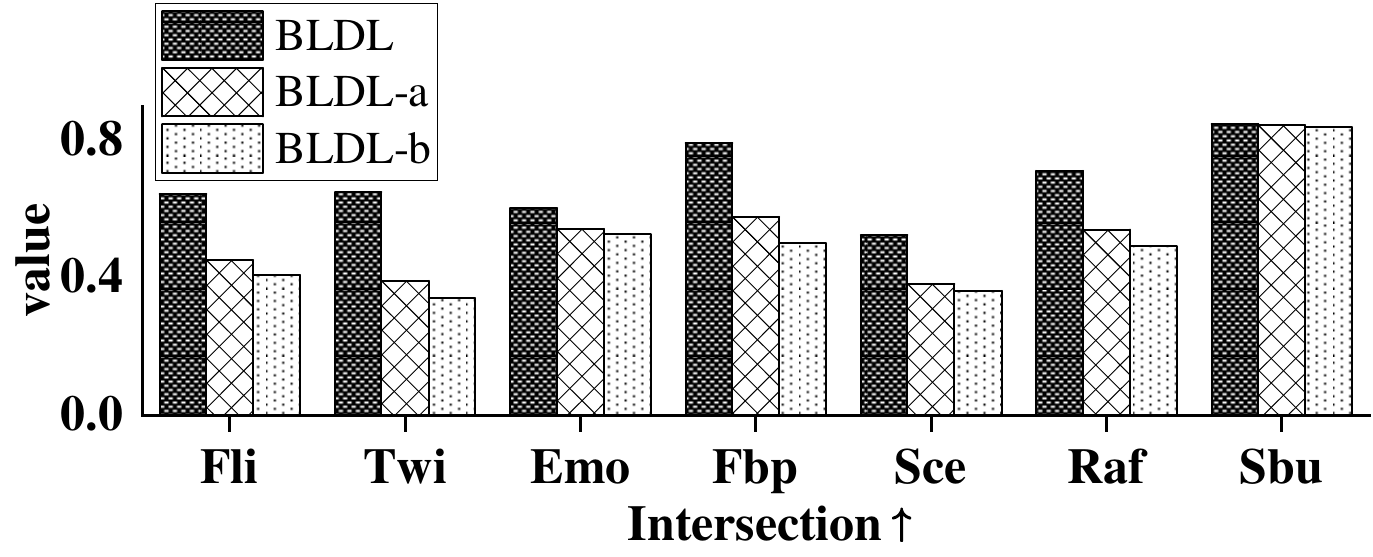}
	}
	\caption{ Ablation results on seven datasets in terms of Chebyshev $\downarrow$, Intersection $\uparrow$.}
	\label{xiaorongkeshihua}
\end{figure*}

\begin{table*}[!h]
	\renewcommand{\arraystretch}{1.2}
	\small
	\centering
	\setlength{\tabcolsep}{1.80mm}
	\begin{tabular}{@{}ccccccc@{}}
		\hline
		BLDL \textit{\textbf{vs}.}& Chebyshev & Clark & KL  & Canberra & Intersection & Cosine \\ \hline
		BLDL-a & $\mathbf {win}$[4.88e-04] & $\mathbf {win}$[3.42e-03] & $\mathbf {win}$ [4.88e-04]& $\mathbf {win}$[4.88e-04] & $\mathbf {win}$[4.88e-04] & $\mathbf {win}$[4.88e-04] \\
		BLDL-b & $\mathbf {win}$[9.77e-04] & $\mathbf {win}$ [9.28e-03]& $\mathbf {win}$[5.37e-03] & $\mathbf {win}$[4.88e-04] & $\mathbf {win}$ [9.77e-04]& $\mathbf {win}$[4.88e-04] \\ \hline
	\end{tabular}
	\caption{The results (Win/Tie/Loss[$p$-value]) of the Wilcoxon signed-rank tests for BLDL against BLDL-a and BLDL-b at a confidence level of 0.05.}
	\label{w-test}
\end{table*}

\begin{figure*}[!h]
	\centering
	\centering
	\subfloat[KL with varying $\alpha$]{
		\includegraphics[scale=0.185, trim={0.4cm 0.5cm 0.5cm 0.5cm}, clip]{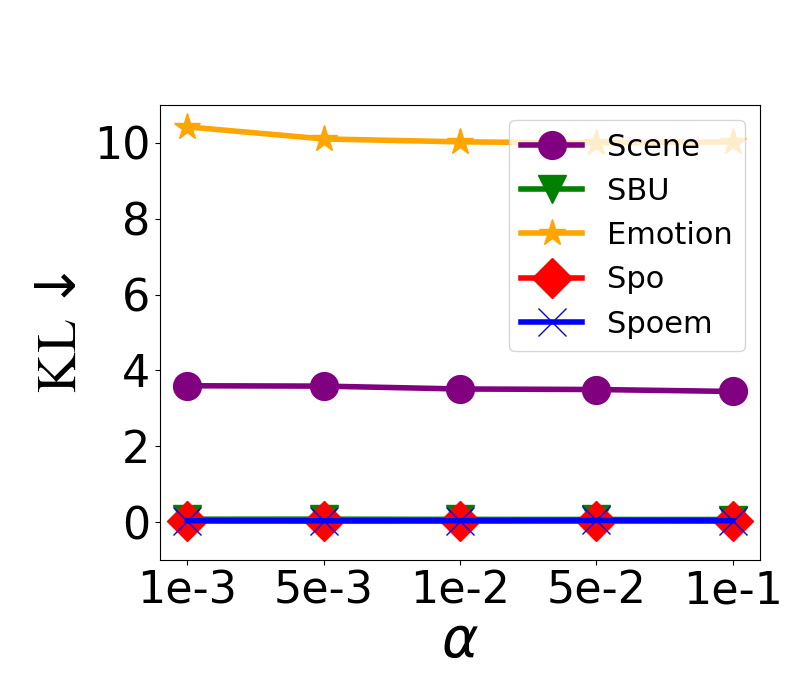}
	}
	\hfil
	\subfloat[Canberra with varying $\beta$]{
		\includegraphics[scale=0.185, trim={0.4cm 0.5cm 0.5cm 0.5cm}, clip]{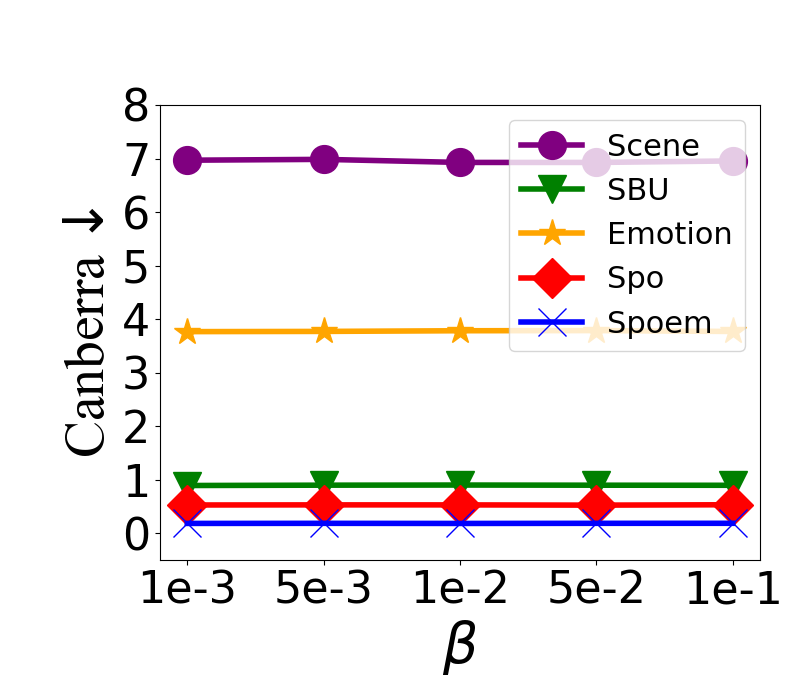}
	}
	\hfil
	\subfloat[Intersection with varying $\eta$]{
		\includegraphics[scale=0.185, trim={0.4cm 0.5cm 0.5cm 0.5cm}, clip]{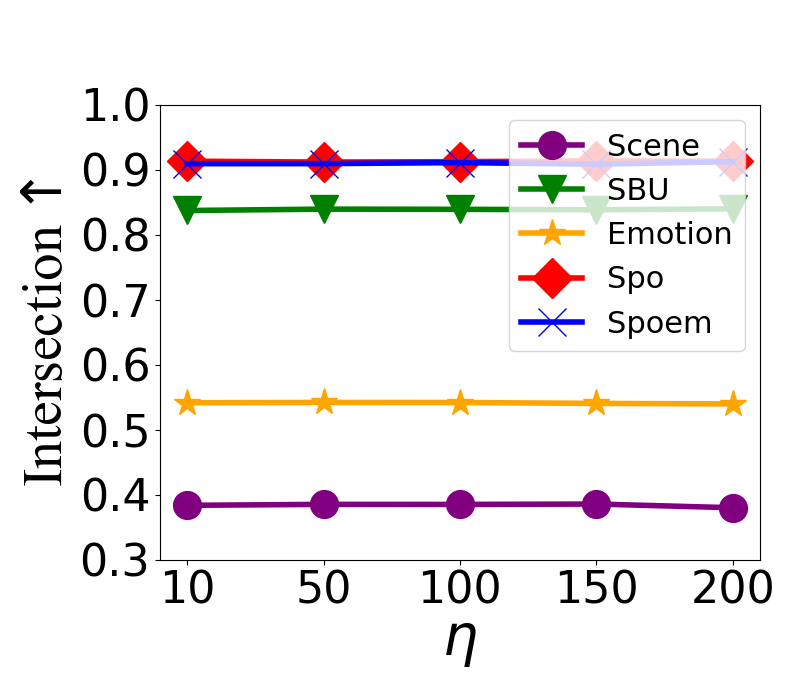}
	}
	\hfil
	\subfloat[Cosine with varying $\lambda_1$]{
		\includegraphics[scale=0.185, trim={0.4cm 0.5cm 0.5cm 0.5cm}, clip]{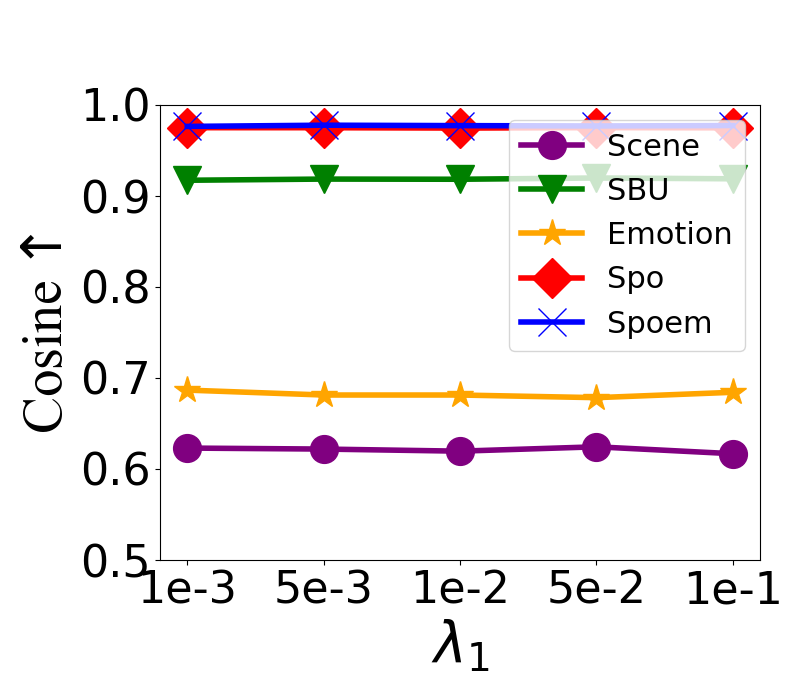}
	}
	\caption{The performance of BLDL with varying parameters in terms of four different metrics on five datasets}
	\label{canshufenxi}
\end{figure*}

\section{Conclusion}
We proposed a unified framework for \textit{joint label distribution recovery and learning}, which performs label distribution recovery while simultaneously training the model. In the recovery phase, our method utilizes biased label distributions and multi-label information to reconstruct the true label distributions effectively. During the learning phase, it exploits multi-label correlations,  rather than in the label distribution space. Both theoretical analysis and extensive experiments demonstrate the effectiveness of the framework, showcasing its ability to mitigate bias and achieve accurate label distribution learning.

\bibliography{example_paper}
\bibliographystyle{icml2025}

\newpage

\section{Appendix}

\subsection{Degradation from biased label distribution to multi-label}

In this paper, we incorporate the multi-label representation derived from the biased label distribution as supplementary information to aid in recovering the true label distribution. We simulate the human annotation process for multi-label images to derive the multi-label matrix from the label distribution. The specific steps are as follows:
\begin{enumerate}
	\item \textbf{Initialization:} Select the most descriptive label \(y_j\) for each instance \(x\), forming an initial set of relevant labels \(\mathcal{Y}^+\).
	\item \textbf{Coverage Calculation:} Compute the cumulative description degree \(H = \sum_{y_j \in \mathcal{Y}^+} d_x^{y_j}\).
	\item \textbf{Threshold Expansion:} Expand \(\mathcal{Y}^+\) iteratively by adding labels maximizing \(d_x^{y_k}\) until \(H > T\), with \(T\) as a predefined threshold.
\end{enumerate}

By executing these steps, we obtain the multi-label matrix \(\hat{\mathbf{L}}\), derived from the biased label distribution.

\subsection{Proof of Algorithm Convergence}

Before the proof, we present here the ADMM updating procedures solving the proposed BLDL model:
\begin{equation*}
	\left\{
	\begin{aligned}
		\mathbf{W}^{k+1} &= \mathbf{A}_1^k\cdot (\mathbf{A}_2^k)^{-1};\\
		\mathbf{O}^{k+1} &= (\mathbf{A}_3^k)^{-1} \cdot \mathbf{A}_4^k;\\
		\mathbf{D}^{k+1} &= \mathbf{A}_5^k\cdot (\mathbf{A}_6^k)^{-1};\\
		\mathbf{Z}^{k+1} &= \text{SVT}_{{1}/{\rho}^k}(\mathbf{W}^{k+1}\mathbf{X}\mathbf{O}^{k+1} - {\boldsymbol{\Lambda}}^{k}/{\rho}^{k});\\
		\boldsymbol{\Lambda}^{k+1} &= \boldsymbol{\Lambda}^k + \rho^k (\mathbf{Z}^{k+1} - \mathbf{W}^{k+1}\mathbf{X}\mathbf{O}^{k+1});\\
		\rho^{k+1} &= \mu \rho^k;
	\end{aligned}
	\right.
\end{equation*}
where
\begin{align*}
	\mathbf{A}_1^{k} &\triangleq 2\alpha \mathbf{D}^{k}\mathbf{X}^\top + \rho^{k} (\mathbf{Z}^{k} + {\boldsymbol{\Lambda}}^{k}/{\rho}^{k})(\mathbf{O}^{k})^\top\mathbf{X}^\top;\\
	\mathbf{A}_2^k &\triangleq 2\alpha \mathbf{X}\mathbf{X}^\top + \rho^{k} \mathbf{X}\mathbf{O}^{k}(\mathbf{O}^{k})^\top\mathbf{X}^\top + 2\lambda_1 \mathbf{I};\\
	\mathbf{A}_3^k &\triangleq 2\beta \hat{\mathbf{D}}^\top \hat{\mathbf{D}} + 2\gamma (\mathbf{D}^{k})^\top \mathbf{D}^{k} \\
	& \ \ \ \ + \rho^{k} \mathbf{X}^\top(\mathbf{W}^{k+1})^\top \mathbf{W}^{k+1}\mathbf{X} + 2\lambda_2 \mathbf{I};\\
	\mathbf{A}_4^k &\triangleq 2\beta \hat{\mathbf{D}}^\top \hat{\mathbf{L}} + 2\gamma (\mathbf{D}^{k})^\top \hat{\mathbf{L}} \\
	& \ \ \ \ + \rho^{k} \mathbf{X}^\top(\mathbf{W}^{k+1})^\top (\mathbf{Z}^{k} + {\boldsymbol{\Lambda}}^{k}/{\rho}^{k});\\
	\mathbf{A}_5^k &\triangleq 2\alpha \mathbf{W}^{k+1}\mathbf{X} + 2\gamma \hat{\mathbf{L}}(\mathbf{O}^{k+1})^\top + 2\eta \hat{\mathbf{D}};\\
	\mathbf{A}_6^k &\triangleq 2\alpha \mathbf{I} + 2\gamma \mathbf{O}^{k+1}(\mathbf{O}^{k+1})^\top + 2\eta \mathbf{I}.
\end{align*}
Note here the iteration number is clearly marked, which is equal to the form in Section 3 where we omit the specific iteration number due to page limitation.

\begin{proof}
	We first prove the boundness for $\mathbf{\Lambda}^{k+1}$.
	\begin{equation}\label{eq.1}
		\begin{split}
			& \ \ \ \  \|\mathbf{\Lambda}^{k+1}\|_F^2\\
			&\overset{(\romannumeral1)}{=} \|\mathbf{\Lambda}^k + \rho^k(\mathbf{Z}^{k+1}-\mathbf{W}^{k+1}\mathbf{X}\mathbf{O}^{k+1})\|_F^2\\
			&=\frac{1}{(\rho^{k})^2} \|\mathbf{W}^{k+1}\mathbf{X}\mathbf{O}^{k+1}-{\mathbf{\Lambda}^{k}}/{\rho}^k-\mathbf{Z}^{k+1}\|_F^2\\
			&\overset{(\romannumeral2)}{=}\frac{1}{(\rho^{k})^2}
			\|\mathbf{U}^k\mathbf{S}^k(\mathbf{V}^k)^\top
			-\mathbf{U}^k (\mathbf{S}^k-{1}/{\rho^k})_+ (\mathbf{V}^k)^\top\|_F^2\\
			&\overset{(\romannumeral3)}{=}\frac{1}{(\rho^{k})^2}
			\|\mathbf{S}^k-(\mathbf{S}^k-1/{\rho^k})_+\|_F^2\\
			&\overset{(\romannumeral4)}{\leq} \frac{N}{(\rho^{k})^4},
		\end{split}
	\end{equation}
	and thus is bounded since $\rho^k$ is monotone increasing, where $N\triangleq\min\{m,n\}$. Here, $(\romannumeral1)$ is from the updating equation of $\mathbf{\Lambda}^{k+1}$, $(\romannumeral2)$ is from the updating equation of $\mathbf{Z}^{k+1}$ via the singular value soft-thresholding operator such that
	$$
	\mathbf{Z}^{k+1} = \mathbf{U}^k (\mathbf{S}^k-{1}/{\rho^k})_+ (\mathbf{V}^k)^\top,
	$$
	where $\mathbf{U}^k\mathbf{S}^k(\mathbf{V}^k)^\top$ is the singular value decomposition of $\mathbf{W}^{k+1}\mathbf{X}\mathbf{O}^{k+1} - {\boldsymbol{\Lambda}}^{k}/{\rho}^{k}$, $(x)_+\triangleq\max\{x,0\}$, $(\romannumeral3)$ holds since $\mathbf{U}^k,\mathbf{V}^k$ is orthogonal and the Frobenius norm is rotation invariant, and last, $(\romannumeral4)$ holds by $\mathbf{S}^k$'s diagonal and the definition of $(\cdot)_+$.
	
	Note that from one round updating process of $\mathbf{W}^{k+1}, \mathbf{O}^{k+1},\mathbf{D}^{k+1}$ and $\mathbf{Z}^{k+1}$ in alternately minimizing the augmented Lagrangian function, it is obvious that
	\begin{equation}\label{eq.2}
		\begin{split}
			& \ \ \ \ \mathcal{L}(\mathbf{W}^{k+1}, \mathbf{O}^{k+1}, \mathbf{D}^{k+1}, \mathbf{Z}^{k+1}, \mathbf{\Lambda}^k, \rho^k)\\
			&
			\leq \mathcal{L}(\mathbf{W}^{k}, \mathbf{O}^{k}, \mathbf{D}^{k}, \mathbf{Z}^{k}, \mathbf{\Lambda}^k, \rho^k).
		\end{split}
	\end{equation}
	Besides, we have
	\begin{equation}\label{eq.3}
		\begin{split}
			& \ \ \ \ \mathcal{L}(\mathbf{W}^{k}, \mathbf{O}^{k}, \mathbf{D}^{k}, \mathbf{Z}^{k}, \mathbf{\Lambda}^k, \rho^k)\\
			&\overset{(\romannumeral1)}{=} \mathcal{L}(\mathbf{W}^{k}, \mathbf{O}^{k}, \mathbf{D}^{k}, \mathbf{Z}^{k}, \mathbf{\Lambda}^{k-1}, \rho^{k-1})\\
			& \ \ \ \ + \langle \mathbf{\Lambda}^k-\mathbf{\Lambda}^{k-1}, \mathbf{Z}^k-\mathbf{W}^k\mathbf{X}\mathbf{O}^k\rangle\\
			& \ \ \ \ +\frac{\rho^k-\rho^{k-1}}{2}\|\mathbf{Z}^k-\mathbf{W}^k\mathbf{X}\mathbf{O}^k\|_F^2\\
			&\overset{(\romannumeral2)}{=} \mathcal{L}(\mathbf{W}^{k}, \mathbf{O}^{k}, \mathbf{D}^{k}, \mathbf{Z}^{k}, \mathbf{\Lambda}^{k-1}, \rho^{k-1})\\
			& \ \ \ \
			+\langle \mathbf{\Lambda}^k-\mathbf{\Lambda}^{k-1}, (\mathbf{\Lambda}^{k}-\mathbf{\Lambda}^{k-1})/\rho^{k-1}\rangle\\
			& \ \ \ \ +\frac{\rho^k-\rho^{k-1}}{2}\|(\mathbf{\Lambda}^{k}-\mathbf{\Lambda}^{k-1})/\rho^{k-1}\|_F^2\\
			&=\mathcal{L}(\mathbf{W}^{k}, \mathbf{O}^{k}, \mathbf{D}^{k}, \mathbf{Z}^{k}, \mathbf{\Lambda}^{k-1}, \rho^{k-1})\\
			& \ \ \ \ +\frac{\rho^k+\rho^{k-1}}{2(\rho^{k-1})^2}\|\mathbf{\Lambda}^{k}-\mathbf{\Lambda}^{k-1}\|_F^2\\
			& \overset{(\romannumeral3)}{\leq} \mathcal{L}(\mathbf{W}^{k}, \mathbf{O}^{k}, \mathbf{D}^{k}, \mathbf{Z}^{k}, \mathbf{\Lambda}^{k-1}, \rho^{k-1})\\
			& \ \ \ \ +(1+\mu)(1+\mu^4)(\rho^0)^{-5}N/{\mu^{5(k-1)}}.
		\end{split}
	\end{equation}
	Here $(\romannumeral1)$ is directly from the formulation of the augmented Lagrangian function, $(\romannumeral2)$ uses the updating equation of $\mathbf{\Lambda}$ at the $k$-th iteration, and $(\romannumeral3)$ uses $\rho^k=\mu\rho^{k-1}=\cdots=\mu^k \rho^0$ and
	\begin{equation}\label{eq.4}
		\begin{split}
			& \ \ \ \ \|\mathbf{\Lambda}^{k}-\mathbf{\Lambda}^{k-1}\|_F^2\\
			&\leq 2\left(\|\mathbf{\Lambda}^{k}\|_F^2+\|\mathbf{\Lambda}^{k-1}\|_F^2\right)\\
			&\leq 2\left(\frac{N}{(\rho^{k-1})^4}+\frac{N}{(\rho^{k-2})^4}\right)\\
			& =  2(1+\mu^4)(\rho^0)^{-4}N/\mu^{4(k-1)}
		\end{split}
	\end{equation}
	where the second inequality uses \eqref{eq.1}. Then, combine \eqref{eq.2} and \eqref{eq.3}, and repeat from $k+1$ to $0$, we have
	\begin{equation}\label{eq.5}
		\begin{split}
			& \ \ \ \ \mathcal{L}(\mathbf{W}^{k+1}, \mathbf{O}^{k+1}, \mathbf{D}^{k+1}, \mathbf{Z}^{k+1}, \mathbf{\Lambda}^k, \rho^k)\\
			&
			\leq \mathcal{L}(\mathbf{W}^{k}, \mathbf{O}^{k}, \mathbf{D}^{k}, \mathbf{Z}^{k}, \mathbf{\Lambda}^{k-1}, \rho^{k-1})\\
			& \ \ \ \ +(1+\mu)(1+\mu^4)(\rho^0)^{-5}N/{\mu^{5(k-1)}}\\
			&\leq \mathcal{L}(\mathbf{W}^{k-1}, \mathbf{O}^{k-1}, \mathbf{D}^{k-1}, \mathbf{Z}^{k}, \mathbf{\Lambda}^{k-2}, \rho^{k-2})\\
			& \ \ \ \ +(1+\mu)(1+\mu^4)(\rho^0)^{-5}N/{\mu^{5(k-2)}}\\
			& \ \ \ \ +(1+\mu)(1+\mu^4)(\rho^0)^{-5}N/{\mu^{5(k-1)}}\\
			& \leq \ldots\\
			&\leq \mathcal{L}(\mathbf{W}^{1}, \mathbf{O}^{1}, \mathbf{D}^{1}, \mathbf{Z}^{1}, \mathbf{\Lambda}^0, \rho^0)\\
			& \ \ \ \ +(1+\mu)(1+\mu^4)(\rho^0)^{-5}N\sum_{i=0}^{k-1}\frac{1}{\mu^{5i}}\\
			&= \mathcal{L}(\mathbf{W}^{1}, \mathbf{O}^{1}, \mathbf{D}^{1}, \mathbf{Z}^{1}, \mathbf{\Lambda}^0, \rho^0)\\
			& \ \ \ \
			+ (1+\mu)(1+\mu^4)(\rho^0)^{-5}N \frac{1-\frac{1}{\mu^{5(k-1)}}}{1-\frac{1}{\mu^5}}\\
			&\leq \mathcal{L}(\mathbf{W}^{1}, \mathbf{O}^{1}, \{\mathbf{G}_i^{1}, \mathbf{\Lambda}_i^{0}, i\in[k]\}, \rho^{0})\\
			& \ \ \ \
			+ (1+\mu)(1+\mu^4)(\rho^0)^{-5}N\mu^5/(\mu^5-1),
		\end{split}
	\end{equation}
	thus has an upper bound. Again, from the formulation of the augmented Lagrangian function, we have the following equation relationship:
	\begin{equation}\label{eq.6}
		\begin{split}
			& \mathcal{L}(\mathbf{W}^{k+1}, \mathbf{O}^{k+1}, \mathbf{D}^{k+1}, \mathbf{Z}^{k+1}, \mathbf{\Lambda}^k, \rho^k) + \frac{1}{2\rho^k}\|\mathbf{\Lambda}^k\|_F^2\\
			&= \|\mathbf{Z}^{k+1}\|_*
			+ \alpha\|\mathbf{W}^{k+1}\mathbf{X} - \mathbf{D}^{k+1}\|_F^2\\
			& + \beta\|\hat{\mathbf{D}}\mathbf{O}^{k+1} - \hat{\mathbf{L}}\|_F^2
			+ \gamma\|\mathbf{D}^{k+1}\mathbf{O}^{k+1} - \hat{\mathbf{L}}\|_F^2\\
			& + \eta\|\mathbf{D}^{k+1} - \hat{\mathbf{D}}\|_F^2
			+ \lambda_1\|\mathbf{W}^{k+1}\|_F^2
			+ \lambda_2\|\mathbf{O}^{k+1}\|_F^2\\
			& + \frac{\rho^k}{2}\|\mathbf{Z}^{k+1} - \mathbf{W}^{k+1}\mathbf{X}\mathbf{O}^{k+1} + {\mathbf{\Lambda}}^{k}/{\rho}^{k}\|_F^2.
		\end{split}
	\end{equation}
	Since both $\mathcal{L}(\mathbf{W}^{k+1}, \mathbf{O}^{k+1}, \mathbf{D}^{k+1}, \mathbf{Z}^{k+1}, \mathbf{\Lambda}^k, \rho^k)$ and $\|\mathbf{\Lambda}^k\|_F^2$ in the left side of \eqref{eq.6} are bounded, then the right side of \eqref{eq.6} is bounded as well and thus each term in the right side of of \eqref{eq.6} is bounded noting that all terms
	are nonnegative. Therefore, $\mathbf{W}^{k+1}, \mathbf{O}^{k+1}, \mathbf{D}^{k+1}$ and $\mathbf{Z}^{k+1}$ are bounded. Note that the above analysis holds for all iteration step $k$.
	
	We now prove the convergence. First,
	\begin{equation}\label{eq.7}
		\begin{split}
			& \ \ \ \ \lim_{k\rightarrow \infty} \|\mathbf{Z}^{k+1}-\mathbf{W}^{k+1}\mathbf{X}\mathbf{O}^{k+1}\|_F^2\\
			&\overset{(\romannumeral1)}{=}\lim_{k\rightarrow\infty}
			\frac{1}{(\rho^k)^2}\|\mathbf{\Lambda}^{k+1}-\mathbf{\Lambda}^k\|_F^2\\
			&\overset{(\romannumeral2)}{\leq} \lim_{k\rightarrow\infty} \frac{1}{(\rho^k)^2}\cdot2(1+\mu^4)(\rho^0)^{-4}N/\mu^{4k}\\
			&\overset{(\romannumeral3)}{=} \lim_{k\rightarrow\infty}
			2(1+\mu^4)(\rho^0)^{-6}N/\mu^{6k}\\
			&=0,
		\end{split}
	\end{equation}
	where $(\romannumeral1)$ is from the updating equation of $\mathbf{\Lambda}^{k+1}$, $(\romannumeral2)$ uses \eqref{eq.6}, and $(\romannumeral3)$ uses $\rho^k=\mu\rho^{k-1}=\cdots=\mu^t \rho^0$. This also yields that
	\begin{equation}
		\lim_{k\rightarrow\infty}\|\mathbf{\Lambda}^{k+1}-\mathbf{\Lambda}^k\|_F^2=0.
	\end{equation}
	
	We then prove $\lim_{t\rightarrow\infty}\|\mathbf{Z}^{k+1}-\mathbf{Z}^k\|_F=0$. By the formulation of the augmented Lagrangian function, we can get
	\begin{equation}\label{eq.8}
		\begin{split}
			& \ \ \ \ \mathcal{L}(\mathbf{W}^{k+1}, \mathbf{O}^{k+1}, \mathbf{D}^{k+1}, \mathbf{Z}^{k+1}, \mathbf{\Lambda}^k, \rho^k)\\
			& \ \ \ \ -\mathcal{L}(\mathbf{W}^{k+1}, \mathbf{O}^{k+1}, \mathbf{D}^{k+1}, \mathbf{Z}^{k}, \mathbf{\Lambda}^k, \rho^k)\\
			&=\|\mathbf{Z}^{k+1}\|_*-\|\mathbf{Z}^k\|_*\\
			& \ \ \ \ + \frac{\rho^k}{2}\left(\|\mathbf{Z}^{k+1}\|_F^2-\|\mathbf{Z}^k\|_F^2\right)\\
			& \ \ \ \ + \langle \mathbf{\Lambda}^k - \rho^k\mathbf{W}^{k+1}\mathbf{X} \mathbf{O}^{k+1}, \mathbf{Z}^{k+1} -\mathbf{Z}^{k}\rangle.
		\end{split}
	\end{equation}
	Suppose that $\mathbf{G}^{k+1}\in\partial \|\mathbf{Z}^{k+1}\|_*$, then the subgradient property gives
	\begin{equation}\label{eq.9}
		\|\mathbf{Z}^{k+1}\|_*-\|\mathbf{Z}^k\|_*
		\leq \langle \mathbf{G}^{k+1}, \mathbf{Z}^{k+1}-\mathbf{Z}^{k}\rangle.
	\end{equation}
	Then,
	\begin{equation}\label{eq.10}
		\begin{split}
			& \ \ \ \ \mathcal{L}(\mathbf{W}^{k+1}, \mathbf{O}^{k+1}, \mathbf{D}^{k+1}, \mathbf{Z}^{k+1}, \mathbf{\Lambda}^k, \rho^k)\\
			& \ \ \ \ -\mathcal{L}(\mathbf{W}^{k+1}, \mathbf{O}^{k+1}, \mathbf{D}^{k+1}, \mathbf{Z}^{k}, \mathbf{\Lambda}^k, \rho^k)\\
			&\leq \frac{\rho^k}{2}\left(\|\mathbf{Z}^{k+1}\|_F^2-\|\mathbf{Z}^k\|_F^2\right)\\
			& \ \ \ \ + \langle \mathbf{G}^{k+1} + \mathbf{\Lambda}^k - \rho^k\mathbf{W}^{k+1}\mathbf{X} \mathbf{O}^{k+1}, \mathbf{Z}^{k+1} -\mathbf{Z}^{k}\rangle.
		\end{split}
	\end{equation}
	By the optimality condition in updating $Z^{k+1}$, then
	\begin{equation*}\label{eq.11}
		\mathbf{G}^{k+1} + \rho^k(\mathbf{Z}^{k+1} -\mathbf{W}^{k+1}\mathbf{X} \mathbf{O}^{k+1} + {\mathbf{\Lambda}^k}/{\rho^k}) = 0.
	\end{equation*}
	Plugging backing to \eqref{eq.10}, we get
	\begin{equation}\label{eq.12}
		\begin{split}
			& \ \ \ \ \mathcal{L}(\mathbf{W}^{k+1}, \mathbf{O}^{k+1}, \mathbf{D}^{k+1}, \mathbf{Z}^{k+1}, \mathbf{\Lambda}^k, \rho^k)\\
			& \ \ \ \ -\mathcal{L}(\mathbf{W}^{k+1}, \mathbf{O}^{k+1}, \mathbf{D}^{k+1}, \mathbf{Z}^{k}, \mathbf{\Lambda}^k, \rho^k)\\
			&\leq \frac{\rho^k}{2}\left(\|\mathbf{Z}^{k+1}\|_F^2-\|\mathbf{Z}^k\|_F^2\right)\\
			& \ \ \ \ + \langle -\rho^k\mathbf{Z}^{k+1}, \mathbf{Z}^{k+1} -\mathbf{Z}^{k}\rangle\\
			&=-\frac{\rho^k}{2}\left(\|\mathbf{Z}^{k+1}\|_F^2
			+ \|\mathbf{Z}^{k}\|_F^2 - 2\langle \mathbf{Z}^{k+1}, \mathbf{Z}^{k}\rangle\right)\\
			&=-\frac{\rho^k}{2} \|\mathbf{Z}^{k+1}-\mathbf{Z}^{k}\|_F^2.
		\end{split}
	\end{equation}
	Note that from one round updating process of $\mathbf{W}^{k+1}, \mathbf{O}^{k+1},\mathbf{D}^{k+1}$ and $\mathbf{Z}^{k+1}$ in alternately minimizing the augmented Lagrangian function, it is easy to get that
	\begin{equation}\label{eq.13}
		\begin{split}
			& \ \ \ \ \mathcal{L}(\mathbf{W}^{k+1}, \mathbf{O}^{k+1}, \mathbf{D}^{k+1}, \mathbf{Z}^{k+1}, \mathbf{\Lambda}^k, \rho^k)\\
			&\leq \mathcal{L}(\mathbf{W}^{k+1}, \mathbf{O}^{k+1}, \mathbf{D}^{k+1}, \mathbf{Z}^{k}, \mathbf{\Lambda}^k, \rho^k)\\
			&\leq \mathcal{L}(\mathbf{W}^{k+1}, \mathbf{O}^{k+1}, \mathbf{D}^{k}, \mathbf{Z}^{k}, \mathbf{\Lambda}^k, \rho^k)\\
			&\leq \mathcal{L}(\mathbf{W}^{k+1}, \mathbf{O}^{k}, \mathbf{D}^{k}, \mathbf{Z}^{k}, \mathbf{\Lambda}^k, \rho^k)\\
			&\leq \mathcal{L}(\mathbf{W}^{k}, \mathbf{O}^{k}, \mathbf{D}^{k}, \mathbf{Z}^{k}, \mathbf{\Lambda}^k, \rho^k).
		\end{split}
	\end{equation}
	Combine \eqref{eq.12} and \eqref{eq.13}, we have
	\begin{equation}\label{eq.14}
		\begin{split}
			& \ \ \ \ \mathcal{L}(\mathbf{W}^{k+1}, \mathbf{O}^{k+1}, \mathbf{D}^{k+1}, \mathbf{Z}^{k+1}, \mathbf{\Lambda}^k, \rho^k)\\
			& \ \ \ \ -\mathcal{L}(\mathbf{W}^{k}, \mathbf{O}^{k}, \mathbf{D}^{k}, \mathbf{Z}^{k}, \mathbf{\Lambda}^k, \rho^k)\\
			&\leq \mathcal{L}(\mathbf{W}^{k+1}, \mathbf{O}^{k+1}, \mathbf{D}^{k+1}, \mathbf{Z}^{k+1}, \mathbf{\Lambda}^k, \rho^k)\\
			& \ \ \ \ -\mathcal{L}(\mathbf{W}^{k+1}, \mathbf{O}^{k+1}, \mathbf{D}^{k+1}, \mathbf{Z}^{k}, \mathbf{\Lambda}^k, \rho^k)\\
			&\leq-\frac{\rho^k}{2} \|\mathbf{Z}^{k+1}-\mathbf{Z}^{k}\|_F^2
		\end{split}
	\end{equation}
	and then
	\begin{equation}\label{eq.15}
		\begin{split}
			& \ \ \ \ \mathcal{L}(\mathbf{W}^{k+1}, \mathbf{O}^{k+1}, \mathbf{D}^{k+1}, \mathbf{Z}^{k+1}, \mathbf{\Lambda}^k, \rho^k)\\
			& \ \ \ \ + \frac{\rho^k}{2} \|\mathbf{Z}^{k+1}-\mathbf{Z}^{k}\|_F^2\\
			& \leq \mathcal{L}(\mathbf{W}^{k}, \mathbf{O}^{k}, \mathbf{D}^{k}, \mathbf{Z}^{k}, \mathbf{\Lambda}^k, \rho^k)\\
			&\leq \mathcal{L}(\mathbf{W}^{k}, \mathbf{O}^{k}, \mathbf{D}^{k}, \mathbf{Z}^{k}, \mathbf{\Lambda}^{k-1}, \rho^{k-1})\\
			& \ \ \ \ +(1+\mu)(1+\mu^4)(\rho^0)^{-5}N/{\mu^{5(k-1)}},
		\end{split}
	\end{equation}
	where the last inequality uses \eqref{eq.3}. Therefore, summing the left and right side of \eqref{eq.15} from $k=1$ to $k=K$ for any sufficiently large $K>0$, we have
	\begin{equation}\label{eq.16}
		\begin{split}
			& \ \ \ \ \sum_{k=1}^K\mathcal{L}(\mathbf{W}^{k+1}, \mathbf{O}^{k+1}, \mathbf{D}^{k+1}, \mathbf{Z}^{k+1}, \mathbf{\Lambda}^k, \rho^k)\\
			& \ \ \ \ + \sum_{k=1}^K\frac{\rho^k}{2}\|\mathbf{Z}^{k+1}-\mathbf{Z}^{k}\|_F^2\\
			& \leq \sum_{k=1}^K\mathcal{L}(\mathbf{W}^{k}, \mathbf{O}^{k}, \mathbf{D}^{k}, \mathbf{Z}^{k}, \mathbf{\Lambda}^k, \rho^k)\\
			&\leq \sum_{k=1}^K\mathcal{L}(\mathbf{W}^{k}, \mathbf{O}^{k}, \mathbf{D}^{k}, \mathbf{Z}^{k}, \mathbf{\Lambda}^{k-1}, \rho^{k-1})\\
			& \ \ \ \ +\sum_{k=1}^K(1+\mu)(1+\mu^4)(\rho^0)^{-5}N/{\mu^{5(k-1)}},
		\end{split}
	\end{equation}
	and then we have
	\begin{equation}\label{eq.16}
		\begin{split}
			& \ \ \ \ \mathcal{L}(\mathbf{W}^{K+1}, \mathbf{O}^{K+1}, \mathbf{D}^{K+1}, \mathbf{Z}^{K+1}, \mathbf{\Lambda}^K, \rho^K)\\
			& \ \ \ \ + \sum_{k=1}^K\frac{\rho^k}{2}\|\mathbf{Z}^{k+1}-\mathbf{Z}^{k}\|_F^2\\
			&\leq \mathcal{L}(\mathbf{W}^{1}, \mathbf{O}^{1}, \mathbf{D}^{1}, \mathbf{Z}^{1}, \mathbf{\Lambda}^0, \rho^0)\\
			& \ \ \ \ +\sum_{t=1}^T\sum_{k=1}^K(1+\mu)(1+\mu^4)(\rho^0)^{-5}N/{\mu^{5(k-1)}}\\
			&\leq \mathcal{L}(\mathbf{W}^{1}, \mathbf{O}^{1}, \mathbf{D}^{1}, \mathbf{Z}^{1}, \mathbf{\Lambda}^0, \rho^0)\\
			& \ \ \ \
			+ (1+\mu)(1+\mu^4)(\rho^0)^{-5}N\mu^5/(\mu^5-1),
		\end{split}
	\end{equation}
	where the last inequality uses the same argument in \eqref{eq.5}. Now letting $K\rightarrow \infty$, since $\mathcal{L}(\mathbf{W}^{k+1}, \mathbf{O}^{k+1}, \mathbf{D}^{k+1}, \mathbf{Z}^{k+1}, \mathbf{\Lambda}^k, \rho^k)$ is bounded for any $k\geq 0$ (see \eqref{eq.5}), then we can get that
	\begin{equation}\label{eq.17}
		\lim_{K\rightarrow \infty}\sum_{k=1}^K\frac{\rho^k}{2}\|\mathbf{Z}^{k+1}-\mathbf{Z}^{k}\|_F^2 < \infty.
	\end{equation}
	Noting that $\|\mathbf{Z}^{k+1}-\mathbf{Z}^{k}\|_F^2$ is bounded and nonnegative, therefore
	\begin{equation}\label{eq.18}
		\lim_{k\rightarrow\infty} \|\mathbf{Z}^{k+1}-\mathbf{Z}^{k}\|_F^2 = 0.
	\end{equation}
	Further, by using the triangle inequality, we have
	\begin{equation}\label{eq.19}
		\begin{split}
			& \ \ \ \ \lim_{k\rightarrow\infty} \|\mathbf{W}^{k+1}\mathbf{X}\mathbf{O}^{k+1}
			-\mathbf{W}^{k}\mathbf{X}\mathbf{O}^{k}\|_F\\
			&\leq \lim_{k\rightarrow\infty} \|\mathbf{W}^{k+1}\mathbf{X}\mathbf{O}^{k+1} -\mathbf{Z}^{k+1}\|_F\\
			& \ \ \ \ +\lim_{k\rightarrow\infty} \|\mathbf{Z}^{k+1}-\mathbf{Z}^{k}\|_F^2\\
			& \ \ \ \ +\lim_{k\rightarrow\infty} \|\mathbf{W}^{k}\mathbf{X}\mathbf{O}^{k}-\mathbf{Z}^{k}\|_F\\
			&=0,
		\end{split}
	\end{equation}
	where the last equation uses \eqref{eq.7} and \eqref{eq.18}. Note that $\mathbf{X}$ is fixed and nonzero, and $\mathbf{W}^{k}, \mathbf{O}^t$ are bounded, thus
	\begin{equation}\label{eq.20}
		\lim_{k\rightarrow\infty} \|\mathbf{W}^{k+1}-\mathbf{W}^k\|_F = 0; \
		\lim_{k\rightarrow\infty} \|\mathbf{O}^{k+1}-\mathbf{O}^k\|_F = 0.
	\end{equation}
	
	Last, following the similar argument, we prove $\lim_{t\rightarrow\infty}\|\mathbf{D}^{k+1}-\mathbf{D}^k\|_F=0$. By the formulation of the augmented Lagrangian function, we have
	\begin{equation}\label{eq.21}
		\begin{split}
			& \ \ \ \ \mathcal{L}(\mathbf{W}^{k+1}, \mathbf{O}^{k+1}, \mathbf{D}^{k+1}, \mathbf{Z}^{k}, \mathbf{\Lambda}^k, \rho^k)\\
			& -\mathcal{L}(\mathbf{W}^{k+1}, \mathbf{O}^{k+1}, \mathbf{D}^{k}, \mathbf{Z}^{k}, \mathbf{\Lambda}^k, \rho^k)\\
			&=(\alpha+\eta)\cdot \left(\|\mathbf{D}^{k+1}\|_F^2-\|\mathbf{D}^{k}\|_F^2\right)\\
			& + \gamma \cdot \left(\|\mathbf{D}^{k+1}\mathbf{O}^{t+1}\|_F^2-\|\mathbf{D}^{k}\mathbf{O}^{t+1}\|_F^2\right)\\
			&
			-2\langle\mathbf{D}^{k+1}-\mathbf{D}^{k}, \alpha\mathbf{W}^{k+1}\mathbf{X}+\gamma\hat{\mathbf{L}}(\mathbf{O}^{k+1})^\top+\eta\hat{D}\rangle.
		\end{split}
	\end{equation}
	Using the updating of $\mathbf{D}^{k+1}$, i.e.,
	\begin{equation*}\label{eq.22}
		\begin{aligned}
			\mathbf{D}^{k+1}
			&= \arg\min_{\mathbf{D}} \alpha\|\mathbf{W}^{k+1}\mathbf{X} - \mathbf{D}\|_F^2 \\
			& \; \; + \gamma\|\mathbf{D}\mathbf{O}^{k+1} - \hat{\mathbf{L}}\|_F^2
			+ \eta\|\mathbf{D} - \hat{\mathbf{D}}\|_F^2.
		\end{aligned}
	\end{equation*}
	By the first-order optimality, there has
	\begin{equation*}\label{eq.23}
		\begin{aligned}
			& -2\alpha(\mathbf{W}^{k+1}\mathbf{X} - \mathbf{D}^{k+1})\\
			& \ \ \ \ +2\gamma(\mathbf{D}^{k+1}\mathbf{O}^{k+1} - \hat{\mathbf{L}})(\mathbf{O}^{k+1})^\top\\
			& \ \ \ \ +2\eta(\mathbf{D}^{k+1}\mathbf{O}^{k+1} - \hat{\mathbf{L}})=0,
		\end{aligned}
	\end{equation*}
	which means that
	\begin{equation*}\label{eq.24}
		\begin{aligned}
			& \ \ \ \ \alpha\mathbf{W}^{k+1}\mathbf{X}+\gamma\hat{\mathbf{L}}(\mathbf{O}^{k+1})^\top+\eta\hat{D}\\
			&=\alpha\mathbf{D}^{k+1}+\gamma\mathbf{D}^{k+1}\mathbf{O}^{k+1}(\mathbf{O}^{k+1})^\top
			+\eta\hat{\mathbf{D}}.
		\end{aligned}
	\end{equation*}
	Plugging backing to \eqref{eq.21}, we get
	\begin{equation}\label{eq.25}
		\begin{split}
			& \ \ \ \ \mathcal{L}(\mathbf{W}^{k+1}, \mathbf{O}^{k+1}, \mathbf{D}^{k+1}, \mathbf{Z}^{k}, \mathbf{\Lambda}^k, \rho^k)\\
			& -\mathcal{L}(\mathbf{W}^{k+1}, \mathbf{O}^{k+1}, \mathbf{D}^{k}, \mathbf{Z}^{k}, \mathbf{\Lambda}^k, \rho^k)\\
			&=(\alpha+\eta)\cdot \left(\|\mathbf{D}^{k+1}\|_F^2-\|\mathbf{D}^{k}\|_F^2\right)\\
			& + \gamma \cdot \left(\|\mathbf{D}^{k+1}\mathbf{O}^{t+1}\|_F^2-\|\mathbf{D}^{k}\mathbf{O}^{t+1}\|_F^2\right)\\
			&
			-2\langle\mathbf{D}^{k+1}-\mathbf{D}^{k}, \\
			& \ \ \ \ \ \alpha\mathbf{D}^{k+1}+\gamma\mathbf{D}^{k+1}\mathbf{O}^{k+1}(\mathbf{O}^{k+1})^\top
			+\eta\hat{\mathbf{D}}\rangle\\
			&=-(\alpha+\eta)\|\mathbf{D}^{k+1}-\mathbf{D}^{k}\|_F^2\\
			& \ \ \ \ -\gamma\|\mathbf{D}^{k+1}\mathbf{O}^{t+1}-\mathbf{D}^{k}\mathbf{O}^{t+1}\|_F^2.
		\end{split}
	\end{equation}
	Therefore,
	\begin{equation}\label{eq.26}
		\begin{split}
			& \ \ \ \ \mathcal{L}(\mathbf{W}^{k+1}, \mathbf{O}^{k+1}, \mathbf{D}^{k+1}, \mathbf{Z}^{k}, \mathbf{\Lambda}^k, \rho^k)\\
			& \ \ \ \ +(\alpha+\eta)\|\mathbf{D}^{k+1}-\mathbf{D}^{k}\|_F^2\\
			& \ \ \ \ +\gamma\|\mathbf{D}^{k+1}\mathbf{O}^{t+1}-\mathbf{D}^{k}\mathbf{O}^{t+1}\|_F^2\\
			& \leq \mathcal{L}(\mathbf{W}^{k+1}, \mathbf{O}^{k+1}, \mathbf{D}^{k}, \mathbf{Z}^{k}, \mathbf{\Lambda}^k, \rho^k)\\
			& \overset{(\romannumeral1)}{\leq} \mathcal{L}(\mathbf{W}^{k}, \mathbf{O}^{k}, \mathbf{D}^{k}, \mathbf{Z}^{k}, \mathbf{\Lambda}^k, \rho^k)\\
			& \overset{(\romannumeral2)}{\leq} \mathcal{L}(\mathbf{W}^{k}, \mathbf{O}^{k}, \mathbf{D}^{k}, \mathbf{Z}^{k}, \mathbf{\Lambda}^{k-1}, \rho^{k-1})\\
			& \ \ \ \ +(1+\mu)(1+\mu^4)(\rho^0)^{-5}N/{\mu^{5(k-1)}}\\
			& \overset{(\romannumeral3)}{\leq} \mathcal{L}(\mathbf{W}^{k}, \mathbf{O}^{k}, \mathbf{D}^{k}, \mathbf{Z}^{k-1}, \mathbf{\Lambda}^{k-1}, \rho^{k-1})\\
			& \ \ \ \ +(1+\mu)(1+\mu^4)(\rho^0)^{-5}N/{\mu^{5(k-1)}},
		\end{split}
	\end{equation}
	where $(\romannumeral1)$ uses \eqref{eq.13}, $(\romannumeral2)$ uses \eqref{eq.9}, and $(\romannumeral3)$ uses \eqref{eq.13} again. Now summing the left and right side of \eqref{eq.26} from $k=1$ to $k=K$ for any sufficiently large $K>0$, and following the same argument as \eqref{eq.16} and \eqref{eq.17}, we can get
	\begin{equation}
		\begin{split}
			& \ \ \ \ \mathcal{L}(\mathbf{W}^{K+1}, \mathbf{O}^{K+1}, \mathbf{D}^{K+1}, \mathbf{Z}^{K}, \mathbf{\Lambda}^K, \rho^K)\\
			& \ \ \ \ + \sum_{k=1}^K(\alpha+\eta)\|\mathbf{D}^{k+1}-\mathbf{D}^{k}\|_F^2\\
			& \ \ \ \ +\sum_{k=1}^K\gamma\|\mathbf{D}^{k+1}\mathbf{O}^{t+1}-\mathbf{D}^{k}\mathbf{O}^{t+1}\|_F^2\\
			&\leq \mathcal{L}(\mathbf{W}^{1}, \mathbf{O}^{1}, \mathbf{D}^{1}, \mathbf{Z}^{0}, \mathbf{\Lambda}^0, \rho^0)\\
			& \ \ \ \
			+ (1+\mu)(1+\mu^4)(\rho^0)^{-5}N\mu^5/(\mu^5-1).
		\end{split}
	\end{equation}
	Since $\mathcal{L}(\mathbf{W}^{k+1}, \mathbf{O}^{k+1}, \mathbf{D}^{k+1}, \mathbf{Z}^{k}, \mathbf{\Lambda}^k, \rho^k)$ is bounded for any $k\geq 0$, letting $K\rightarrow \infty$, then we can get that
	\begin{equation}
		\lim_{K\rightarrow \infty}\sum_{k=1}^K(\alpha+\eta)\|\mathbf{D}^{k+1}-\mathbf{D}^{k}\|_F^2 < \infty,
	\end{equation}
	and thus
	\begin{equation}
		\lim_{k\rightarrow\infty} \|\mathbf{D}^{k+1}-\mathbf{D}^{k}\|_F^2 = 0,
	\end{equation}
	since $\alpha,\eta>0$.  The proof is completed.
	
\end{proof}

\end{document}